\documentclass[lettersize,journal]{IEEEtran}
\usepackage{amsmath,amsfonts}
\usepackage{array}
\usepackage[caption=false,font=normalsize,labelfont=sf,textfont=sf]{subfig}
\usepackage{textcomp}
\usepackage{stfloats}
\usepackage{url}
\usepackage{verbatim}
\usepackage{graphicx}
\usepackage{color}
\usepackage{cite}
\usepackage[ruled,linesnumbered]{algorithm2e}

\usepackage{times}
\usepackage{soul}
\usepackage{url}
\usepackage[hidelinks]{hyperref}
\usepackage[utf8]{inputenc}
\usepackage[small]{caption}
\usepackage{graphicx}
\usepackage{amssymb}
\usepackage{booktabs}
\usepackage{multirow,makecell}
\usepackage{ulem}
\usepackage{diagbox}
\begin{document}

\title{Towards Optimal Customized Architecture for Heterogeneous Federated Learning with Contrastive Cloud-Edge Model Decoupling}

\author{
\IEEEauthorblockN{
Xingyan Chen,~\IEEEmembership{Member,~IEEE,}
Tian Du,
Mu Wang$^\dag$,
Tiancheng Gu,
Yu Zhao,~\IEEEmembership{Member,~IEEE,}
Gang Kou$^\dag$,
Changqiao Xu,~\IEEEmembership{Senior Member,~IEEE,}
Dapeng Oliver Wu~\IEEEmembership{Fellow,~IEEE,}
    \IEEEcompsocitemizethanks{
		\IEEEcompsocthanksitem Manuscript received xx; revised xx; accepted xx. Date of publication; date of current version. This work was supported in part by the National Natural Science Foundation of China under Grant 62225105, 62101301, 62302400.
        \IEEEcompsocthanksitem	X. Chen, T. Du, T. Gu, Y. Zhao, and G. Kou are with Financial Intelligence and Financial Engineering Key Laboratory of Sichuan Province, Institute of Digital Economy and Interdisciplinary Science Innovation, School of Computer and Artificial Intelligence, Southwestern University of Finance and Economics, Chengdu 611130, P. R. China. 
		E-mail: \{xychen, zhaoyu, kougang\}@swufe.edu.cn, \{dtian.nora, elegy12138\}@gmail.com. 
		\IEEEcompsocthanksitem	M. Wang and C. Xu are with the State Key Laboratory
		of Networking and Switching Technology, Beijing University of Posts and
		Telecommunications, Beijing 100876, P. R. China.
		E-mail: \{wangmu, cqxu\}@bupt.edu.cn.
        \IEEEcompsocthanksitem	D. Oliver Wu is with the Data Engineering at the Department of Computer Science, City University of Hong Kong. 
		E-mail: dapengwu@cityu.edu.hk.
		\IEEEcompsocthanksitem $^\dag$Corresponding author: Mu Wang and Gang Kou.
    }
  }
}
\maketitle

\begin{abstract}
Federated learning, as a promising distributed learning paradigm, enables collaborative training of a global model across multiple network edge clients without the need for central data collecting.
However, the heterogeneity of edge data distribution drags the model towards the local minima, which can be distant from the global optimum. Such heterogeneity often leads to slow convergence and substantial communication overhead.
To address these issues, we propose a novel federated learning framework called {\tt FedCMD}, a model decoupling tailored to the Cloud-edge supported federated learning that separates deep neural networks into a body for capturing shared representations in Cloud and a personalized head for migrating data heterogeneity.
Our motivation is that, by the deep investigation of the performance of selecting different neural network layers as the personalized head, we found rigidly assigning the last layer as the personalized head in current studies is not always optimal. Instead, it is necessary to dynamically select the personalized layer that maximizes the training performance by taking the representation difference between neighbor layers into account. To find the optimal personalized layer, we utilize the low-dimensional representation of each layer to contrast feature distribution transfer and introduce a Wasserstein-based layer selection method, aimed at identifying the best-match layer for personalization.
Additionally, a weighted global aggregation algorithm is proposed based on the selected personalized layer for the practical application of {\tt FedCMD}.
Extensive experiments on ten benchmarks demonstrate the efficiency and superior performance of our solution compared with nine state-of-the-art solutions.
All code and results are available at \url{https://github.com/elegy112138/FedCMD}.

\end{abstract}


\section{Introduction}
\IEEEPARstart{I}{mpressive} capacity magnification in wireless communication technologies has catalyzed a significant expansion in the capabilities of mobile devices, wearables, and other internet-connected devices. These devices boast advanced hardware and software, enabling the support of various applications and the generation of extensive volumes of contextual and user-specific data. Due to privacy concerns and the enhanced computing and storage capacities at the network edge, there is an increasing preference to store data locally on edge devices~\cite{Xu2023age,zhang2023digital}. This shift facilitates the execution of the model training on these devices using locally available data, with only sporadic communication with a central parameter server. This decentralized methodology, known as federated learning~\cite{mcmahan2017communication}, is a stark departure from traditional model training, which typically relies on the central aggregation of raw data.

\begin{figure}[!t]
	\begin{center}
	\includegraphics[width=\linewidth]{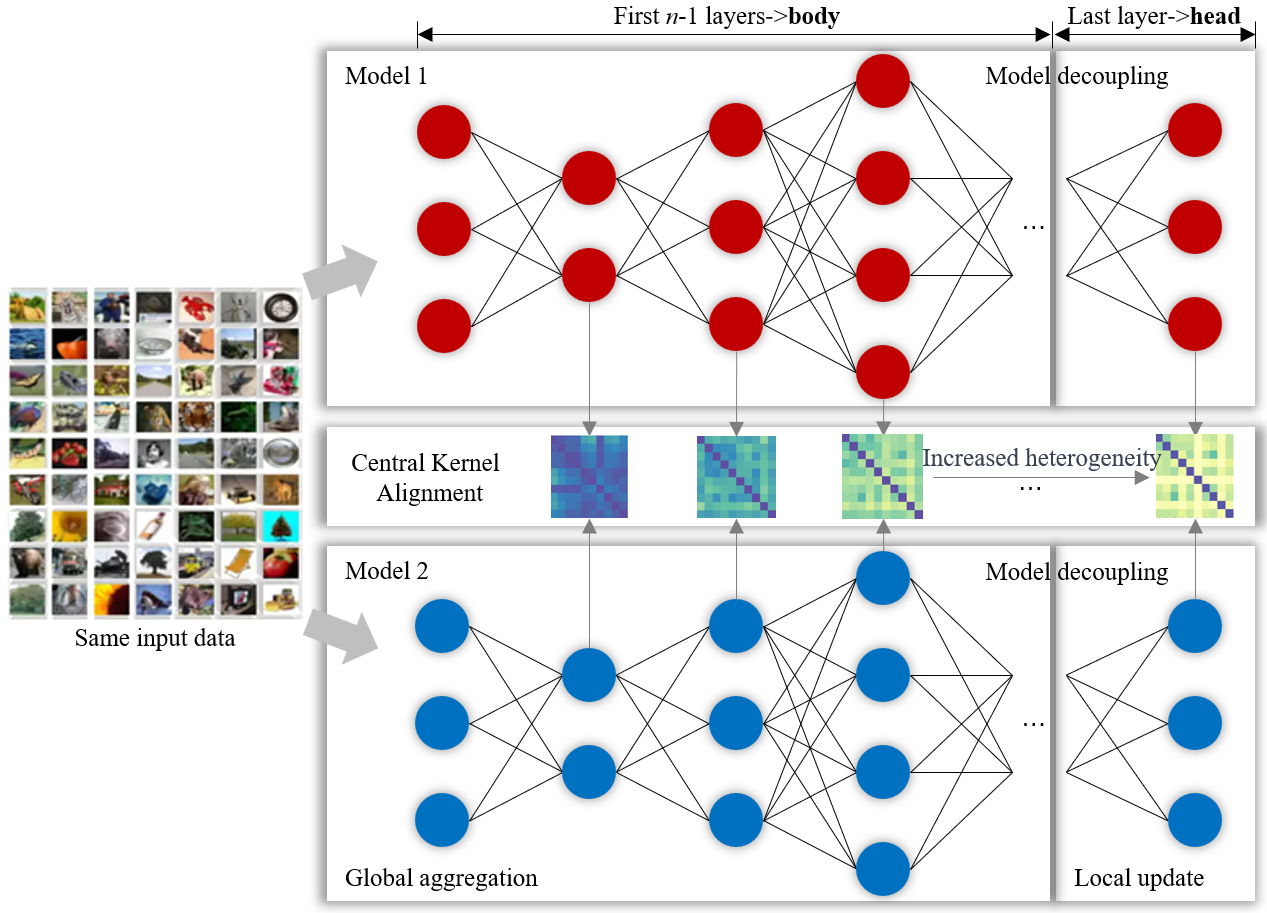}
	\end{center}
	\begin{center}
		\vspace{-1em}
		\caption{Model decoupling based on Central Kernel Alignment.}
		\label{CKA}
	\end{center}
	\vspace{-3em}
\end{figure}
One of the most significant challenges in federated learning, as highlighted by recent research, is the heterogeneity of data distribution (non-independent and identically distributed, non-IID) across a multitude of client devices. 
In this scenario, where potentially thousands of devices participate, the heterogeneity can significantly impact the convergence of the model training and cause increased communication costs.
This challenge is especially acute when training large-scale neural network models, such as transformer-based models, which are surprisingly difficult to train~\cite{lu2021soft}.

Addressing the adverse effects of data heterogeneity is a critical area of research, with numerous recent studies dedicated to developing solutions.
These include approaches like knowledge distillation~\cite{chen2023metafed,zhang2023towards,yao2024fedgkd}, loss regularization~\cite{li2020federated,t2020personalized,li2021ditto}, similarity aggregation~\cite{long2023multi,sattler2020clustered,yang2023personalized}, and model decoupling~\cite{collins2023exploiting,oh2021fedbabu,tan2023pfedsim}.
Among these, model decoupling, a strategy that has gained prominence recently, stands out for its notable capability to significantly reduce communication overhead and enhance performance by strategically segmenting deep neural networks into a shared pre-output \textbf{body} and a distinct personalized \textbf{head}~\cite{oh2021fedbabu,collins2023exploiting}. 
As shown in Fig. \ref{CKA}, assuming the local model is an {\it n}-layer neural network, prior work~\cite{luo2021fear,collins2023exploiting,oh2021fedbabu,liang2020think,tan2023pfedsim} conventionally designated the first ({\it n}-1) layers as the \textbf{body} and the last layer as the personalized \textbf{head}. 
In this paradigm, the pre-output body parameters undergo global aggregation, while the personalized layer is only updated locally. This approach primarily aims to mitigate the adverse effects of heterogeneous data on shared partial parameters to the central server, thus enhancing the personalized performance~\cite{tan2023pfedsim}.

However, this last-layer-as-head decoupling, rooted in Central Kernel Alignment (CKA)~\cite{collins2023exploiting,oh2021fedbabu,tan2023pfedsim}, does not demonstrate optimal guarantees in theory.
CKA is used as the metric to evaluate the similarity of outputs of the same layer from two identical-structure models with the same input data. 
In scenarios where network edges with heterogeneous data distributions, the locally trained models process information layer-by-layer with the degree of heterogeneity of each layer's outputs increasing throughout this progression.
Based on this, current methods contend that the output of the final layer, exhibiting the highest heterogeneity, is most suitable for the personalized layer.
This perspective fails to consider that the heterogeneity noted in the output of the last layer is not isolated, but rather the cumulative effect of all previous layers.

Moving from theoretical considerations to practical implications, we conducted extended experiments across ten real-world datasets with three different neural architectures also revealing that the last layer might not always be the best selection. 
These observations motivate us to rethink the existing last-layer-as-head design.
Thus, we conducted an in-depth exploration for model decoupling and found three statistical discoveries based on the experiment results: (1) Selecting the last layer as the personalized layer is not the best option in most cases; (2) While multiple layers can be chosen as personalized head, optimal performance emerges from selecting the most appropriate single layer; (3) Consistency in choosing the same personalized layer across different clients is essential for fast and stable convergence of the model training.
Building on the experimental findings, we also discovered that the criteria for selecting the personalized layer should consider both the input and output of each layer, rather than focusing only on the output as is typical in CKA-based solutions.

In this context, there is a need to find a new method that can knowledgeize the heterogeneity of data distribution across different users to facilitate the personalized layer selection.
Thus, we first propose a layer-wise knowledgeization method to capture the low-dimensional feature distribution of the intermediate layers.
Furthermore, a novel contrastive personalized layer selection mechanism based on Wasserstein distance~\cite{ludger1985} is proposed to identify the best-match layer.
This method involves a layer-by-layer comparison to identify transfer in low-dimensional representational distributions of each layer. Subsequently, selecting the layer where the feature distribution transfer is minimized as the personalized layer.
Extensive experimental testing has shown that choosing the layer with the minimal Wasserstein distance relative to the local data distribution as the personalized layer yields the best performance.
Further, we developed a novel contrastive Cloud-edge model decoupling framework named {\tt FedCMD} by dividing the Cloud-Edge federated learning into two main phases: the personalized layer selection phase and the \textcolor{black}{heterogeneous federated learning phase}.
We also provided two piratical algorithms by selecting the personalized layer with the contrastive layer selection to migrate data heterogeneity at the edges and share the other layers' parameters to the Cloud for collaborative learning.
To the best of our knowledge, {\tt FedCMD} represents the first attempt to introduce an alternative to the CKA standard for personalized layer selection, proposing the use of a non-final layer as the personalized layer in model decoupling.
The main contributions of this work are as follows:
\begin{itemize}
    \item Introducing a new distribution distance metric called feature distribution transfer for personalized layer selection, based on three findings discovered through extensive experimentation results on real-world datasets.
    \item Proposing a novel heterogeneous federated learning framework, termed {\tt FedCMD} with two main phases based on the feature distribution shift distance. To the best of our knowledge, our work is the first exploration in facilitating personalized layer selection by quantifying the heterogeneity in their data distributions.
    \item Providing two practical algorithms for achieving personalized layer selection and weighted global aggregation across the Cloud-Edge federated learning system.
    \item Conducting comprehensive experiments, 
    validating our solution's performance superiority on ten datasets compared to \textcolor{black}{nine} other SOTA solutions.
\end{itemize}

The related works are reviewed in section II.
Section III presents the motivation.
Section IV provides the system model and preliminaries.
Section V gives the details of the contrastive layer selection mechanism and {\tt FedCMD} framework.
Section VI introduces the algorithm design and implementation.
In section VII, we evaluate our solution in five main parts with \textcolor{black}{nine} state-of-the-art solutions.
Finally, section VIII concludes the paper and outlines directions for future work.

\section{Related Works}

\subsection{Heterogeneous Federated Learning}
Federated Learning such as FedAvg~\cite{mcmahan2017communication} is a method designed for privacy-preserving data sharing while allowing the training of models across distributed devices.
However, traditional federated learning, mainly developed for scenarios with independent and identically distributed (iid) data distribution, struggles with slow convergence rates when applied to heterogeneous data distribution. 

To address the challenge, heterogeneous federated learning has been proposed which encompasses several key aspects: (1) \textbf{Knowledge distillation}~\cite{zhang2023towards,chen2023metafed,yao2024fedgkd} refines the exchange of information between global and local models, enhancing model generalization capabilities. (2) \textbf{Regularization terms} in the loss function~\cite{li2020federated,t2020personalized,li2021ditto} are employed to correct biases towards the global model, improving the robustness and universality of the global model. (3) \textbf{Similarity aggregation}, such as clustering~\cite{sattler2020clustered,long2023multi,yang2023personalized} and multi-task learning~\cite{sattler2020clustered,li2021ditto,he2022spreadgnn}, enable more efficient information sharing among similar clients or tasks. Clustering groups like clients to learn individual models, whereas multi-task learning handles several client models simultaneously without forming distinct clusters. However, these methods often come with inherent drawbacks, such as relying on additional public datasets~\cite{zhang2023towards,chen2023metafed,yao2024fedgkd}, increased computational complexity~\cite{li2020federated,t2020personalized,li2021ditto}, and the risk of data leakage caused by additional parameter exposure~\cite{sattler2020clustered,li2021ditto,he2022spreadgnn,long2023multi,yang2023personalized}. One of the currently prominent approaches is (4) \textbf{Model decoupling}, which segregates models into shared \textbf{body} and personalized \textbf{head}. The body's layers are shared among all clients, whereas the head is tailored based on each client's data, striking a balance between global performance and individual client requirements.

\begin{table*}
\caption{Personalized layer performances of models with one fc and classifier across 10 datasets.}
\centering
\footnotesize 
\begin{tabular}{lcccccccccc}
 \hline
    Configuration & CIFAR100 & CIFAR10 & CINIC10 & EMNIST & FMNIST & M.N.A & M.N.C & MNIST & SVHN & T. ImageNet \\
    \hline
    fc1 & 40.346 & \textbf{85.265} & \textbf{80.27} & 92.968 & 96.548 & 94.192 & 92.897 & 98.827 & 92.990 & 22.345 \\
    classifier & \textbf{43.647} & 84.556 & 78.792 & \textbf{93.635} & \textbf{96.750} & \textbf{96.017} & \textbf{94.376} & \textbf{99.336} & \textbf{93.873} & \textbf{28.201} \\
    \hline
\end{tabular}
\vspace{-.5em}
\label{tab:swapped_motivation1}
\end{table*}

\begin{table*}
\caption{Personalized layer performances of models with two fc and classifier across 10 datasets.}
\centering
\footnotesize 
\begin{tabular}{lcccccccccc}
 \hline
    Configuration & CIFAR100 & CIFAR10 & CINIC10 & EMNIST & FMNIST & M.N.A & M.N.C & MNIST & SVHN & T. ImageNet \\
    \hline
    fc1 & 42.698 & 86.263 & 80.500 & 93.188 & 96.420 & 95.096 & 85.742 & 98.956 & 93.302 & 21.740 \\
    fc2 & \textbf{47.541} & \textbf{86.940} & \textbf{80.575} & \textbf{93.994} & \textbf{96.599} & \textbf{96.487} & \textbf{94.172} & \textbf{99.421} & \textbf{94.654} & \textbf{31.056} \\
    classifier & 42.925 & 84.282 & 78.983 & 93.554 & 96.403 & 95.588 & 94.029 & 99.363 & 94.240 & 28.061 \\
    \hline
\end{tabular}
\vspace{-.5em}
\label{tab:swapped_motivation2}
\end{table*}

\begin{table*}
\caption{Personalized layer performances of models with three fc and classifier across 10 datasets.}
\centering
\footnotesize 
\begin{tabular}{lcccccccccc}
 \hline
    Configuration & CIFAR100 & CIFAR10 & CINIC10 & EMNIST & FMNIST & M.N.A & M.N.C & MNIST & SVHN & T. ImageNet \\
    \hline
    fc1 & 39.340 & 86.110 & 80.230 & 93.096 & 96.633 & 94.807 & 92.603 & 98.874 & 93.732 & 20.487 \\
    fc2 & \textbf{46.895} & \textbf{87.088} & \textbf{80.677} & 93.825 & \textbf{96.701} & \textbf{96.645} & \textbf{94.623} & \textbf{99.381} & \textbf{95.042} & 30.135 \\
    fc3 & 44.975 & 84.305 & 79.060 & \textbf{94.000} & 96.366 & 95.985 & 94.147 & 99.337 & 94.669 & \textbf{30.732} \\
    classifier & 39.723 & 83.512 & 78.282 & 87.407 & 90.933 & 95.514 & 92.837 & 98.261 & 89.523 & 27.444 \\
    \hline
\end{tabular}
\vspace{-.5em}
\label{tab:swapped_motivation3}
\end{table*}

\begin{table*}
    \caption{Performances comparison of multiple layers and single layer as the personalized layer across 10 datasets.}
    \centering
    \begin{tabular}{lcccccccccc}
    \hline
    Configuration & CIFAR100 & CIFAR10 & CINIC10 & EMNIST & FMNIST & M.N.A & M.N.C & MNIST & SVHN & T.ImageNet \\
    \hline
    fc2 & 
    \textbf{47.541} & \textbf{86.940} & \textbf{80.575} & \textbf{93.994} & \textbf{96.599} & \textbf{96.487} & \textbf{94.172} & \textbf{99.421} & \textbf{94.654} & \textbf{31.056} \\
    fc1 \& fc2 & 40.669 & 84.491 & 80.186 & 93.027 & 96.286 & 94.080 & 92.693 & 98.671 & 92.818 & 21.786 \\
    fc1 \& classifier & 40.735 & 85.265 & 80.260 & 93.001 & 96.407 & 94.721 & 93.068 & 98.859 & 93.144 & 21.716 \\
    fc2 \& classifier & 43.313 & 84.811 & 79.741 & 93.850 & 96.450 & 95.550 & 93.499 & 99.263 & 94.132 & 28.887 \\
    fc1 \& fc2 \& classifier & 38.458 & 83.475 & 78.582 & 92.828 & 96.164 & 93.393 & 91.893 & 98.477 & 92.453 & 21.635 \\
    \hline
    \end{tabular}
    \vspace{-1.5em}
    \label{tab:motivation2}
\end{table*}


\subsection{Model Decoupling}
Model decoupling is a critical approach that distinctly separates the fundamental and personalized layers of the model.  Several pivotal methodologies, including FedRep~\cite{collins2023exploiting}, FedPer~\cite{arivazhagan2019federated}, and FedPav~\cite{wei2023personalized} achieve this by splitting client models into fundamental and personalized layers. In these models, the foundational layers are sent to the server for global aggregation, while the personalized layers are updated locally at each client. For instance, FedBN~\cite{li2021fedbn} and FedAP~\cite{lu2022personalized} use the batch normalization layer as the personalized component, keeping it local, with other layers participating in global aggregation to reduce feature drift. Notably, FedAP employs Wasserstein distance to assess the batch normalization layer’s data, aiding in setting aggregation weights. In contrast, FedBabu~\cite{oh2021fedbabu} and pFedSim~\cite{tan2023pfedsim} classify the final layer (classifier) as the \textbf{head} and earlier layers as the \textbf{body}. FedBabu uses a fixed \textbf{head} initialized randomly, while pFedSim updates the \textbf{head} locally and uses weighted global aggregation based on classifier distance. However, these methods are too rigid and inflexible in personalized layer selection, and cannot adapt the layer selection based on the specific situation. Compared to existing methods, our work presents several notable distinctions and advantages. Firstly, we conducted an extensive exploration of the current last-layer-as-head decoupling design and gave three findings in Section \ref{Motivation}. Secondly, we proposed a novel contrastive layer selection mechanism based on Wasserstein distance. Lastly, leveraging the personalized layer selection outcomes, we further devised a weighted global aggregation approach to enhance personalized performance.

\section{Motivation}\label{Motivation}
We present motivation from three perspectives: (1) identifying the best layer for personalization, (2) evaluating whether multiple layers could enhance performance, and (3) \textcolor{black}{determining if the optimal personalized layers vary across different clients with heterogeneous data distributions.}



\begin{figure*}[htbp]
    \centering
    \begin{minipage}[b]{1\linewidth}
        \centering
        \includegraphics[width=\linewidth]{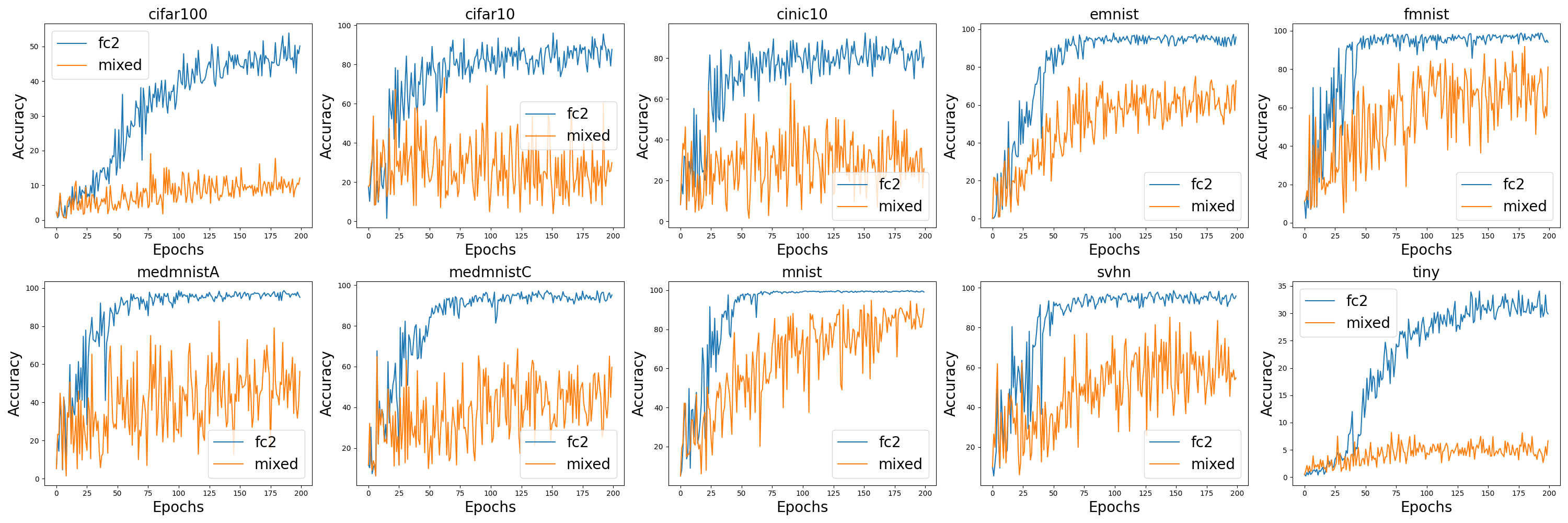}
    \end{minipage}
    \vspace{-1.5em}
    \caption{Performance comparison of the mixed personalized layer approach and the fc2 personalized layer method across ten datasets.}
    \vspace{-1.5em}
    \label{fig:motivation3}
\end{figure*}

\subsection{Performance comparison with different personalized layer}
Studies like FedRep~\cite{collins2023exploiting}, FedBadu~\cite{oh2021fedbabu} and pFedSim~\cite{tan2023pfedsim}, which adopt Central Kernel Alignment theory, typically define the first ({\it n}-1) layers of a neural network as the \textbf{body} and the last layer as personalized \textbf{head}. 
However, our experiments with three different neural architectures across ten datasets (more details, please refer to Appendix \ref{appendixA}) and as illustrated in Tables \ref{tab:swapped_motivation1}--\ref{tab:swapped_motivation3}, we observed that the pre-output fully connected (fc) layer performs better than the last layer (classifier) as the personalized layer in most cases.
We tested three versions of the LeNet5~\cite{lecun1998gradient} model, each with a different number of fc layers (2, 3, and 4 layers), across ten standard datasets. In the 2-layer model as shown in Table \ref{tab:swapped_motivation1}, the classifier layer showed better results than the first fc layer (\textit{fc1}) when used as the personalized layer. For the 3-layer LeNet5 as shown in Table \ref{tab:swapped_motivation2}, the second fc layer (\textit{fc2}) yielded the highest performance when used as the personalized layer across all datasets. With the 4-layer model as shown in Table \ref{tab:swapped_motivation3}, both the \textit{fc2} and third fc layers (\textit{fc3}) demonstrated superior performance compared to using the classifier as the personalized layer.
The results indicate that the classifier is not always the best choice.
This finding across these configurations highlights the need to develop a new mechanism for personalized layer selection.


\subsection{Multi-layer personalization evaluation}
Recent studies such as FedTSDP~\cite{zhu2023federated} and PCCFED~\cite{su2023novel}, are exploring the use of multiple fc layers as the personalized layer. To investigate whether employing a combination of fc layers as the personalized layer could enhance performance, we conducted experiments with the 3-layer LeNet5 model. As shown in Table. \ref{tab:motivation2}, we compared the performance of using just the \textit{fc2} layer as the personalized layer against four combinations: (1) \textit{fc1}+\textit{fc2}, (2) \textit{fc1}+\textit{classifier}, (3) \textit{fc2}+\textit{classifier}, and (4) \textit{fc1}+\textit{fc2}+\textit{classifier}. Our findings reveal that the performance of the single \textit{fc2} layer always surpassed the combinations. \textcolor{black}{This suggests that selecting multiple layers as the personalization does not lead to performance improvement.} 

\subsection{Heterogeneity of clients in personalized layer selection}
Our experiment results also revealed that despite all clients selecting \textit{fc2} as their personalized layer, the performance of their local models varied. This observation prompted us to explore the possibility that clients with heterogeneous data distributions may benefit from having different personalized layers.
We conducted experiments where clients that showed poor performance with \textit{fc2} as their personalized layer were reassigned to use the \textit{classifier} instead. This adjustment was implemented to evaluate the effectiveness of a mixed personalized layer case on the model training.
However, this mixed solution did not enhance model performance and led to slow convergence in ten different datasets, as shown in Fig.~\ref{fig:motivation3}.
Based on the above results, our solution will focus on the design where all clients choose one common personalized layer. 

\begin{table}[!t]
\renewcommand{\arraystretch}{1}
\caption{Mathematical Notations}
\label{notations}
\vspace{-0.5em}
\centering
\begin{tabular}{| c || l |}
\hline
\textbf{Symbol} & \multicolumn{1}{| c |}{\textbf{Description}}\\
\hline
$\mathcal{G}(\mathcal{V},\mathcal{L})$ & the topology graph of the Cloud-edge system. \\
\hline
$\mathcal{V},\mathcal{L}$ & Set of nodes and links for the system. \\
\hline
$\mathcal{C}$ & The set of the edge clients. \\
\hline
$N$ & The number of edge clients. \\
\hline
$\mathcal{K} $ & The set of communication round. \\
\hline
$\mathcal{D}_i$ & The local dataset with $i$-th client. \\
\hline
${\mathcal{X}_i,\mathcal{Y}_i}$ & The set of input data and label for $i$-th client respectively. \\
\hline
$n_i$ & The data size of the $i$-th client. \\
\hline
$n$ & The total number of samples across all devices. \\
\hline
$\mathcal{C}^k$ & The set of clients participates in FL in round $k$. \\ 
\hline
$n^k$ & The number of participated clients in round $k$. \\
\hline
$\gamma$ & The join ratio of the edge clients in FL. \\
\hline
$\theta_G^k$ & The global model parameter in round $k$. \\
\hline
$\theta_i^k$ & The $i$-th client model parameter in round $k$. \\
\hline
$\omega$ & The parameters with the common layers. \\
\hline
$\phi$ & The parameters with the personalized layers. \\
\hline
$\text{W}_p(\mu, \nu)$ & The \textit{p}-order Wasserstein Distance of $\mu$ and $\nu$. \\
\hline
$z^{o_l}$ & The feature distribution of the output of $l$-th layer. \\
\hline
$\rho$ & The division ratio of the two main phases.\\
\hline
$K_p$ & The number of communication rounds in the first phase. \\
\hline
$\mathcal{S}$ & The candidate list of the personalized layer. \\
\hline
$l^*$ & The selected personalized layer. \\
\hline
$\Phi_{ij}$ & The similarity matrix of client $i$ and $j$. \\
\hline
\end{tabular}
\vspace{-2em}
\end{table}

\section{Preliminaries}
In this section, we give the preliminaries of the system model, federated Learning workflow with model decoupling, and Wasserstein distance.
For notation, scalars are represented by lowercase italic letters. Vectors and sets are denoted by lowercase bold italics and uppercase italics, respectively. Matrices are indicated with uppercase, italic, and bold fonts. The symbol \(\Delta\) represents the differential operator, while \(|\cdot|\) and \(\|\cdot\|\) are used for the 1-norm and 2-norm, respectively. Table \ref{notations} lists the mathematical notations used in this paper.

\subsection{System Model}
We consider the Cloud-edge system as an undirected graph $\mathcal{G}(\mathcal{V},\mathcal{L})$ where $\mathcal{V}$ and $\mathcal{L}$ are the sets of nodes and links.
Let $\mathcal{C} \subset \mathcal{V}$ as the set of the edge clients, with $N=|\mathcal{C}|$ indicating the number of edge clients.
We consider a time-slotted system where the communication round is represented as $\mathcal{K} = \{1,2,...\}$.
The $i$-th client is associated with a local dataset, denoted as $\mathcal{D}_i$.
Since the data distribution are heterogeneous, every dataset $\mathcal{D}_i=\{\mathcal{X}_i,\mathcal{Y}_i\}=\{x_j^{(i)},y_j^{(i)}\}_{j=1}^{n_i}$ is non-independently and identically distributed, where $\mathcal{X}_i$ and $\mathcal{Y}_i$ represent the set of input data and label for $i$-th client respectively, and $n_i$ represents the data size of the $i$-th client.
${\it n_i}=|\mathcal{D}_i|$, $x_j$ is the $j$-th input data, $y_j$ is the corresponding label. 
The total number of samples across all devices equals $n = \sum_{i \in \mathcal{C}} n_i$ and the total dataset is $\mathcal{D} = \{\mathcal{X},\mathcal{Y}\}=\{\mathcal{D}_i\}_{i\in \mathcal{C}}$.

\subsection{Federated Learning with Model Decoupling}

\paragraph{Federated learning workflow} 
The training process encompasses $K$ communication rounds. 
In each round $k$, a subset of clients participates, defined by $\mathcal{C}^k \subset \mathcal{C}$, and the number of participated clients is denoted as $n^k=|\mathcal{C}^k| = \gamma \cdot |\mathcal{C}|$, where $k \in [0, K]$ and $ \gamma$ is the join ratio. 
We consider the classical procedure of federated learning. Initially, the server distributes the model $\theta_G^k$ to a select group of clients, initializing parameters as $\theta_G^0$. Subsequently, clients receive the model and execute local training with the local datasets $\mathcal{D}_i$. The overall objective of federated learning is expressed as:
\begin{equation}
    \underset{\theta}{\arg \min} \frac{1}{N} \sum_{i\in\mathcal{C}} f_i(\theta),
\end{equation}
where $f_i(\theta)=\mathbb{E}_{(x,y) \sim \mathcal{D}_i} [l(\theta;(x,y))]$, with $l$ denoting the local empirical loss function of $i$-th client. 
After local training, clients send their model parameters $\{\theta_i^k\},i\in \mathcal{C}^k$ back to the server for the global aggregation phase:
\begin{equation}
    \theta_G^k=\sum_{i\in \mathcal{C}^k} \frac{n_i}{n^k} \theta_i^k,
    \label{eq:fedavg_aggregation}
\end{equation}
where $n^k=\sum_{i\in \mathcal{C}^k} n_i$ is the total number of samples of the client participating in round $k$.

\paragraph{Model decoupling} Formally, a model with parameters $\theta$ can be decoupled as $\theta = \omega \circ \phi$, where $\omega$ represents the common layers and $\phi$ represents the personalized layers. Correspondingly, the objective function of federated learning in model decoupling is represented by:
\begin{equation}
\underset{\omega,\{\phi_i\}}{\arg \min} \frac{1}{N} \sum_{i\in\mathcal{C}} f_i(\omega,\phi_i),
\end{equation}
considering $\theta_i = \omega \circ \phi_i$. In this setup, $\omega$ is consistent across all clients, denoting the shared components, whereas $\{\phi_i\}$ is the set of personalized layer parameters uniquely for each of the edge clients, enabling personalization.

\begin{figure*}[htbp]
    \centering
    \begin{minipage}[b]{1\linewidth}
        \centering
        \includegraphics[width=\linewidth]{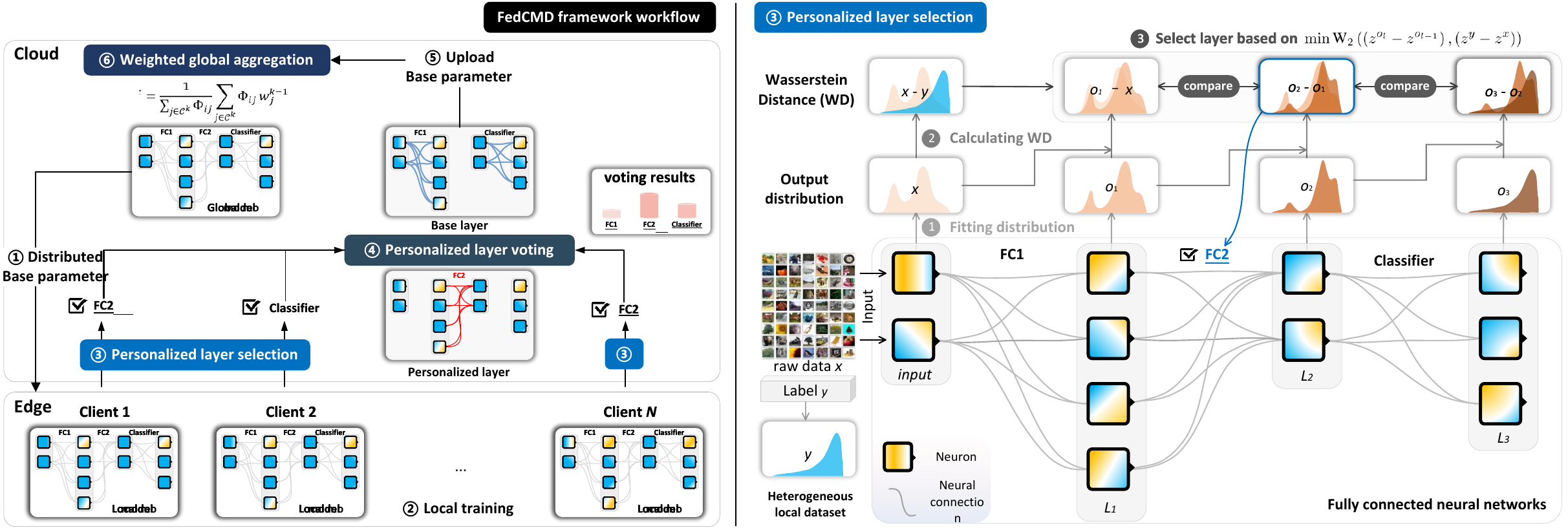}
    \end{minipage}
    \vspace{-1.5em}
    \caption{The framework of {\tt FedCMD} with dynamic personalized layer selection.}
    \vspace{-1.5em}
    \label{fig:framework}
\end{figure*}


\subsection{Wasserstein Distance}
The Wasserstein distance~\cite{ludger1985} is a measure used to quantify the difference between two probability distributions. It represents the minimum effort required to reshape one distribution into another. For distributions \(\mu\) and \(\nu\), the Wasserstein distance is defined as the lowest expected cost necessary for transforming \(\mu\) into \(\nu\). It is mathematically expressed as:
\begin{equation}
     \text{W}_p(\mu, \nu) = \left( \inf_{\gamma \in \Gamma(\mu, \nu)} \int_{U \times V} c(\mu, \nu)^p \, d\gamma(\mu, \nu) \right)^{\frac{1}{p}},
\end{equation}
where, \(U\) and \(V\) represent the spaces of the distributions \(\mu\) and \(\nu\). \(\Gamma(\mu, \nu)\) includes all possible joint distributions that conform to the marginal distributions \(\mu\) and \(\nu\). The term \(c(\mu, \nu)\) is the cost function measuring the distance between elements in \(\mu\) and \(\nu\), and \(p\) indicates the order of the Wasserstein distance. This metric is widely used in areas such as data mining~\cite{bigot2020stat}, image processing~\cite{she2023improve}, and generative modeling~\cite{lee2023}.

\begin{figure}[htbp]
    \centering
    \begin{minipage}[b]{1\linewidth}
        \centering        \includegraphics[width=\linewidth]{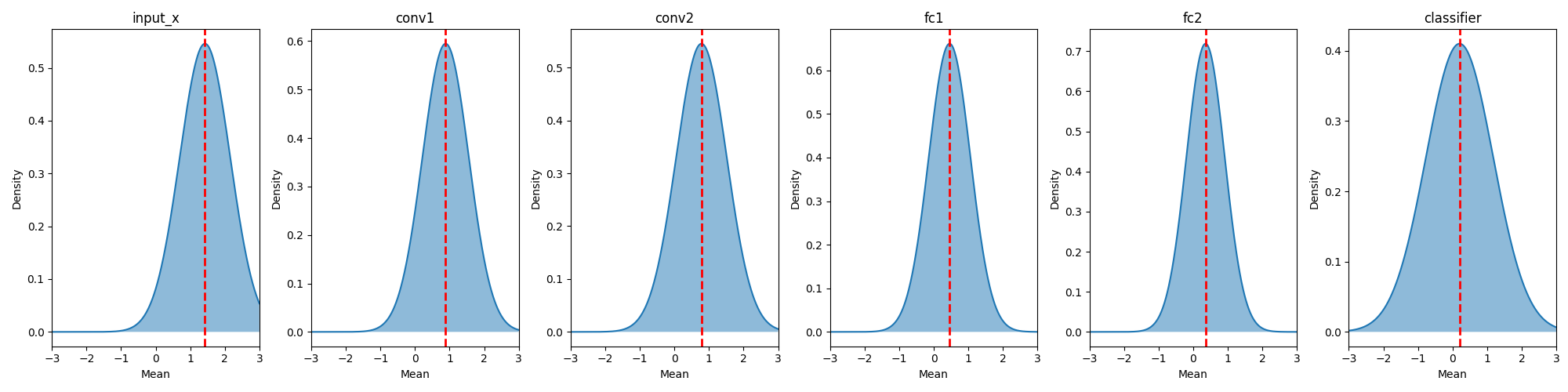}
    \end{minipage}
    \vspace{-1.5em}
    \caption{The feature distribution of different layers for LeNet5~\cite{lecun1998gradient}}
    \vspace{-1.5em}
    \label{fig:fd}
\end{figure}

\section{Framework with contrastive layer selection mechanism}
In this section, we first present an contrastive layer selection mechanism that employs a new distance measurement based on Wasserstein distance. 
Then, we illustrate the {\tt FedCMD} framework as applied within the Cloud-Edge system.

\subsection{Contrastive layer selection mechanism}
Before discussing the contrastive layer selection, we introduce a new metric for distribution distance measurement, \textcolor{black}{named feature distribution transfer}, which is based on the Wasserstein distance, and the formulation is as follows:
\begin{equation}
    s_l = \min \text{W}_2\left(\left(z^{o_l} - z^{o_{l-1}}\right), \left(z^y - z^x\right)\right),
    \label{eq:s_l}
\end{equation}
where $\text{W}_2(\cdot)$ denotes the Wasserstein distance function with $p=2$.
The terms \(o_l\) and \(o_{l-1}\) refer to the outputs of the \(l\)-th and \((l-1)\)-th layers of the neural network, respectively. 
Their associated low-dimensional feature distributions of the \(l\)-th and \((l-1)\)-th layers are denoted by \(z^{o_l}\) and \(z^{o_{l-1}}\).
The variables \(x\) and \(y\) represent the input data sample and its corresponding label in the dataset.
Their associated feature distributions are denoted by \(z^x\) and \(z^y\), which capture the representations of the input and label data according to the data distribution.
\textcolor{black}{Fig. \ref{fig:fd}, shows the Gaussian distribution of each layer's outputs for LeNet5 model in the CIFAR-10 dataset.} From this visualization, it is observable that the feature distributions of the intermediate layers are gradually shifting toward zero-mean distribution (CIFAR-10 is a class-balanced dataset). Furthermore, the extent of this shift, such as \textit{conv2} to \textit{fc1} and \textit{fc1} to \textit{fc2}, is not uniform across different layers.
\textcolor{black}{The feature distribution transfer distance, \(s_l\),} thus quantifies the alignment of the low-dimensional feature transformations of the neural architecture from \(z^{o_l}\) to \(z^{o_{l-1}}\) with the data distribution transfer from inputs \(z^x\) to labels \(z^y\).

The formulation of the feature distribution transfer metric, \( s_l \), is grounded in the principle that intermediate hidden layers of a neural network capture low-dimensional feature information. Represented by \( o_l \) and \( o_{l-1} \), these outputs embody the low-dimensional features extracted by the network. In cases where clients have heterogeneous data distributions, the low-dimensional feature information they extract also varies accordingly. To address this, we represent the low-dimensional feature distributions of the \( l \)-th and \((l-1)\)-th layers as \( z^{o_l} \) and \( z^{o_{l-1}} \) respectively, and for simplicity, we model \( z \) as a Gaussian distribution.

The rationale behind this approach is our belief that the optimal personalized layer should closely align with the unique data distribution characteristics of each client. Since \( z^x \) and \( z^y \) capture the heterogeneous data distribution, we constructed \( s_l \) to evaluate this alignment. This metric quantifies how closely the neural network's feature transformations from \( z^{o_l} \) to \( z^{o_{l-1}} \) match the shifts in data distribution from inputs \( z^x \) to labels \( z^y \), thus guiding the selection of the most suitable personalized layer in alignment with each client's heterogeneous data.

Due to the black-box nature of neural networks, it is challenging to precisely compute the feature distribution transfer at each layer, hence estimation is employed. Leveraging $x$ and $y$ as references, the difference between the distribution distances of the current layer output and the original data $x$, original labels $y$, and the preceding layer output and the original data $x$, original labels $y$ is utilized as an estimate of the current layer's feature distribution transfer. Eq. (\ref{eq:s_l}) can be further rewritten as:
\begin{equation}
    \begin{aligned}
        s_l &= \min \text{W}_2\left(\left(z^{o_l} - z^{o_{l-1}}\right), \left(z^y - z^x\right)\right) \\
        &\approx \min \big|\underbrace{\left(\text{W}_2\left(z^{o_l},z^y\right) - \text{W}_2\left(z^{o_l},z^x\right)\right)}_{\text{Prat A}} - \\ 
        &\underbrace{\left(\text{W}_2\left(z^{o_{l-1}},z^y\right) - \text{W}_2\left(z^{o_{l-1}},z^x\right)\right)}_{\text{Part B}}\big|.
    \end{aligned}
    \label{eq:max_sl}
\end{equation}

The reason for employing the $l_1$ norm is that a negative value for the Wasserstein distance holds no meaningful interpretation. Part A represents the Wasserstein distance between the distribution transfer of the $l$-th layer output and the original data, while Part B represents the Wasserstein distance between the $(l-1)$-th layer output's distribution transfer and the original data, highlighting the difference as an estimate of the $l$-th layer's feature distribution transfer increment.

\subsection{{\tt FedCMD} Framework}
The proposed {\tt FedCMD} framework is illustrated in \textcolor{black}{Fig \ref{fig:framework}}.
We consider the Cloud-edge cooperation system, in which two types of nodes are involved, including one Cloud server, and multiple edge clients.
{\tt FedCMD} contains two major phases including the personalized layer selection phase and the heterogeneous federated learning phase with the Cloud-edge model decoupling.
In \textbf{personalized layer selection phase}, the Cloud-edge system employs the standard federated learning such as {\tt FedAvg} and utilizes the contrastive layer selection mechanism to collaboratively elect the personalized layer. 
Thus, the selection proceeds as follows:
\begin{itemize}
\item \textbf{Global parameters broadcasting}: The central server initializes the global model parameter and distributes them to all edge clients.
\item \textbf{Local model updating}: After receiving the global parameter, each client updates their local model and trains the model using the local data.
\item \textbf{Local Layer Selection}: Based on the criteria outlined in Eq. \eqref{eq:max_sl}, each edge client selects a personalized layer from their model during the training.
\item \textbf{Communication of local gradients}: Edge clients send updates of their local model parameters and their chosen personalized layer back to the Cloud server.
\item \textbf{Cloud-side processing}: The Cloud server applies the standard federated aggregation to update the parameter of the global model and determine the layer selected by the most number of edge clients as the personalized layer.
\end{itemize}
Once this personalized layer is determined, it remains fixed throughout the \textbf{heterogeneous federated learning phase}. The personalized layer's parameters do not participate in global aggregation and are updated locally on the edge client side. In contrast, the other layers of the model parameters are updated through a weighted federated aggregation, the specifics of which will be detailed in the algorithm design section.
The process of the federated learning phase can be outlined as:
\begin{itemize}
\item \textbf{Global parameters broadcasting}: The central Cloud server transmits the global model parameter, excluding the personalized layer, to edge clients.
\item \textbf{Local model updating}: Upon receiving the global parameter, each edge client updates their local model and trains the model using the local data.
\item \textbf{Communication of local gradients}: Edge clients send updates of their local model parameters back to the Cloud server.
\item \textbf{Cloud-side processing}: The Cloud server utilizes the personalized layer parameters to calculate the weight matrix and implements weighted federated aggregation to update the parameters of the body layers.
\end{itemize}


\section{Algorithm Design}
\subsection{{\tt FedCMD} design}
{\tt FedCMD} approach comprises two distinct phases: the personalized layer selection phase and the heterogeneous federated learning phase, involving three key sub-algorithms: Client-side update, Personalized layer selection, and Heterogeneous federated learning.

\begin{algorithm}
    \caption{\textbf{ClientUpdate}($i,\theta_i^k$): // \textit{run on client i}}
    \label{alg:algorithm1}
    \SetKwInput{KwInput}{Input}
    \SetKwInput{KwOutput}{Output}
    \KwInput{The Cloud server paremeters $\theta_i^k$.}
    \KwOutput{The local model parameters $\theta_i^k$ or $(\omega_i^k,\phi_i^k)$ updated with local data.}
    Splits $\theta_i^k \to (\omega_i^k,\phi_i^k)$;
    
    \For{each local epoch}{
        Update $\theta_i^k$ or $(\omega_i^k,\phi_i^k)\leftarrow SGD_i(\omega_i^k,\phi_i^k;\mathcal{D}_i)$ \;}
        \textbf{return} $\theta_i^k$ or $(\omega_i^k,\phi_i^k)$ \;
\end{algorithm}

\begin{algorithm}
    \caption{Personalized Layer Selection}
    \label{alg:algorithm2}
    \SetKwInput{KwInput}{Input} 
    \SetKwInput{KwOutput}{Output} 
    \KwInput{The number of current phase communication round $K_p$, the candidate list $\mathcal{S}$.}
    \KwOutput{The personalized layer $l^*$.}
    The Cloud Server delivers the initialization parameter $\theta_G^0$ to the edge clients;

    Calculate the mean and std of the local data and fit the distribution $z^{x}$ and $z^{y}$;
    
    \For{$k=1,\ldots,K_p$}{
        \textcolor[rgb]{0.41568,0.52549,0.56078}{/* Run on the edge clients */}
    
        $\mathcal{C}^k \leftarrow$ \textit{(random set of $n^k$ clients from $\mathcal{C}$)};
        
        \For{each client $i \in \mathcal{C}^k$ in parallel}{
            Receive $\theta_G^{k-1}\to \theta_i^{k-1}$ from the Cloud server;
        
            $\theta_i^k \leftarrow $ \textbf{ClientUpdate} $(i,\theta_i^{k-1})$;

            \For{each layer output $o_l$}{
                Calculate the mean and std for each layer $l$ and fit the distribution $z^{o_l}$ and $z^{o_{l-1}}$;
                
                Calculate $s_l$ by Eq. (\ref{eq:max_sl}) and add the corresponding layer into candidate list $\mathcal{S}$; }

            Update $\theta_i^k$ to the Cloud server;
            }

        \textcolor[rgb]{0.41568,0.52549,0.56078}{/* Run on the Cloud server */}

        Select the layer that appears most frequently in $\mathcal{S}$ as the personalized layer for $\mathcal{C}^k$, and \textit{recording} it; 
        
        $\theta_G^{k} \leftarrow \sum_{i \in \mathcal{C}^k} \frac{|\mathcal{D}_i|}{|\mathcal{D}|}\theta_i^k$, where $|\mathcal{D}|=\sum_{i\in \mathcal{C}^k}|\mathcal{D}_i|$;

        Broadcast $\theta_G^{k}$ to all edge clients.
        }
        
    Select the most frequent layer in $K_p$ rounds as the final decision of the personalization layer $l^*$.
\end{algorithm}

\paragraph{Client-side updates} For each client $i$, the client model parameters $\theta_i^k$ or $(\omega_i^k,\phi_i^k)$ are updated using local data $\mathcal{D}_i$ through the stochastic gradient descent approach. The pseudocode for this process is illustrated in \textbf{Algorithm \ref{alg:algorithm1}}.

\paragraph{Personalized Layer Selection Phase} 
Let the number of total communication rounds be denoted as $K$, and the number of current phase communication rounds as $K_p$.
For each communication round, utilize Eq. \eqref{eq:max_sl} to compute the distribution distance 
between each layer's $z^{o_l}-z^{o_{l-1}}$ and $z^{y}-z^{x}$. Identify the layer corresponding to the minimum distance, and then, through a voting mechanism, determine the most frequent layer to augment the candidate list $\mathcal{S}$ at round $k$. Upon completion of the final communication round in this stage, select the most recurrent layer from \(S\) as the fixed personalized layer \(l^*\) for the subsequent stage. The aggregation method for this stage follows the classical FedAvg\cite{mcmahan2017communication} approach. See \textbf{Algorithm \ref{alg:algorithm2}} for the detailed process.

\paragraph{Heterogeneous Federated Learning Phase} In the stage of heterogeneous federated learning, the personalized selection outcome layer \(l^*\), derived from the preceding phase, is intended to be employed throughout subsequent \textcolor{black}{$K-K_p$} communication rounds. In this phase, each client personalization layer is set to \(l^*\) and remains unchanged, \(i.e.\), it is updated locally but no other layers are selected. Moreover, based on the results of personalized layer selection, a weighted parameter aggregation is applied to the model update, according to Eq. (\ref{eq:aggregation}) and (\ref{eq:phi}). \textbf{Algorithm \ref{alg:algorithm3}} shows the detailed process.

\subsection{Weighted Global Aggregation} 
Following the personalized-layer-selection outcome $l^*$, the global aggregation is divided into two distinctive parts: 1) Layers preceding the personalized layer primarily emphasize feature extraction, leaning towards generality. Leveraging the sample weighting approach introduced in {\tt FedAvg}~\cite{mcmahan2017communication} as Eq. (\ref{eq:fedavg_aggregation}), these layers undergo an averaging process. Conversely, 2) layers succeeding the personalized layer tend to accentuate heterogeneity, drawing inspiration from {\tt pFedSim}~\cite{tan2023pfedsim}. The aggregation involves utilizing the cosine similarity of personalized layer parameters for weighted averaging, which can be calculated as follows: 
\begin{equation}
\begin{aligned}
\theta_i^k = \phi_i^k\circ \omega_i^k
           = \left\{    
	\begin{aligned}
	& \omega_i^k=\frac{1}{\sum_{j \in \mathcal{C}^k} \Phi_{ij}} \sum_{j \in \mathcal{C}^k} \Phi_{ij}\omega_j^{k-1}, \\
	& \phi_i^k= \phi_i^{k-1},
	\end{aligned}
\right.
\end{aligned}
\label{eq:aggregation}
\end{equation}
where $\Phi_{ij}$ can be further written as Eq (\ref{eq:phi}). Here, we define $\epsilon$ as a small positive value (we set $10^{-8}$ in our experiments) to avoid yielding extreme values.
We denote $\theta_{l^*}(i)$ as the parameters of the \textcolor{black}{selected personalized layer $\phi_i$} for client $i$.
\begin{equation}
\begin{aligned}
    \Phi_{ij} &= \mathrm{cosine}[\theta_{l^*}(i),\theta_{l^*}(j)]\\
    &= \frac{\phi_i\cdot\phi_j}{\parallel \phi_i\parallel\cdot\parallel\phi_j\parallel+\epsilon},
\end{aligned}
    \label{eq:phi}
\end{equation}
where $\Phi_{ij}\in[0,1]$, and a higher value implies that clients $i$ and $j$ are more similar. 
The aggregation approach outlined in this context holds the potential for further extension into more sophisticated forms to tackle challenges across diverse scenarios. {\tt FedCMD} refrains from exchanging any additional information apart from model parameters. Crucially, it preserves the personalized layer parameters, refraining them from undergoing the aggregation process. In contrast to alternative baseline methods, it showcases reduced expenses and a superior degree of privacy safeguarding.

\begin{algorithm}
    \caption{Heterogeneous Federated Learning}
    \label{alg:algorithm3}
    \SetKwInput{KwInput}{Input} 
    \SetKwInput{KwOutput}{Output} 
    \KwInput{The number of total communication round $K$, number of current phase communication round $K_p$, \textcolor{black}{the personalized layer $l^*$}}
    \KwOutput{The personalized parameters of edge clients, \(\{\omega_1^K,\omega_2^K,\ldots,\omega_N^K\}\)}
    Receive $\theta_G^{K_p}$ from previous phase for all edge clients in $\mathcal{C}$ and decouple it into $\omega_G^{K_p},\phi_G^{K_p}$;
    
    \For{$k={K_p}+1,\ldots,K$}{
        \textcolor[rgb]{0.41568,0.52549,0.56078}{/* Run on the edge clients */}
    
        $\mathcal{C}^k \leftarrow$ (\textit{random set of $n^k$ clients}) ;
        
        \For{each client $i\in \mathcal{C}^k$ in parallel}{
            Receive $\omega_G^{k-1}\to \omega_i^{k-1}$ from the Cloud server;
            
            Get $\theta_i^{k-1} \leftarrow \omega_i^{k-1} \circ \phi_i^{k-1}$;
            
            $(\omega_i^k,\phi_i^k) \leftarrow$ \textbf{ClientUpdate}$(i,\theta_i^{k-1})$;

            Update $(\omega_i^k,\phi_i^k)$ to the Cloud server;
            }

        \textcolor[rgb]{0.41568,0.52549,0.56078}{/* Run on the Cloud server */}
        
        \For{two different clients $i,j\in\mathcal{C}^k$}{
            Compute the similarity matrix $\Phi_{ij}$ by Eq. (\ref{eq:phi});
            } 

            Update $\{\omega_i^{k}\}$ based on Eq. \eqref{eq:aggregation} and broadcast ge to the $i$-th edge client;
            }
\end{algorithm}

\subsection{Implementation and Complexity Analysis}
{\tt FedCMD} consists of two crucial stages: personalized layer selection and heterogeneous federated learning. In this subsection, we provide an implementation and complexity overview for each stage based on the algorithm's description.

Firstly, we consider the client-side update, a subprocess utilized in both two phases. Assuming each stochastic gradient descent (SGD) operation is an \(\mathcal{O}(1)\) process. The time complexity of the \textbf{ClientUpdate} algorithm focuses on two factors: the number of local epochs \(E\) and the number of samples \(n_i\) of the client \(i\). Therefore, the time complexity is \(\mathcal{O}(E \cdot n_k)\).

In the edge client workflow of the personalized layer selection phase, the process starts each communication round by selecting a set of edge clients \(\mathcal{C}^k\) from the total pool \(\mathcal{C}\), which has a complexity of \(\mathcal{O}(|\mathcal{C}|)\). 
Then, the local trains on the chosen edge clients \(\mathcal{C}^k\) are parallel performed using the \textbf{ClientUpdate}. For a model comprising \(L\) layers, we assume the complexity of each edge client to calculate and fit the Gaussian distribution \(\{z^{o_l}\}\) to be \(\mathcal{O}(L)\), and the Wasserstein distance computation as per Eq. (\ref{eq:max_sl}) to be \(\mathcal{O}(1)\). 
The complexity of this step is \(\mathcal{O}(|\mathcal{C}^k| + E \cdot n_k + L + 1)\). 
Given that \(L\) is significantly smaller than both \(\mathcal{C}^k\) and \(n_k\), we simplify the time complexity to \(\mathcal{O}(|\mathcal{C}^k| + E \cdot n_k)\).
On the Cloud server side, the time complexity of both each round's personalized layer selection voting and server parameter updating is \(\mathcal{O}(|\mathcal{C}^k|)\). 
Algorithm \ref{alg:algorithm2} repeats the above process for \(K_p\) rounds, leading to the overall time complexity of \(\mathcal{O}(K_p \cdot (|\mathcal{C}^k| + E \cdot n_k))\).

Finally, we analyze the time complexity of the second phase.
On the edge client side, involves the model decoupling, and \textbf{ClientUpdate}. We consider the time complexity of model decoupling as \(\mathcal{O}(1)\) and \textbf{ClientUpdate} is \(\mathcal{O}\left(|\mathcal{C}^k| + E \cdot n_k\right)\).
The time complexity on the client side is $\mathcal{O}\left(|\mathcal{C}^k| + E \cdot n_k\right)$. Moreover, there is a need to compute the cosine similarity matrix $\Phi$ between any two clients on the Cloud server.
Assuming the time complexity of Eq. \eqref{eq:phi} is \(\mathcal{O}(1)\).
The steps from 12 to 14 of \textbf{Algorithm \ref{alg:algorithm3}} is $\mathcal{O}\left(\left|\mathcal{C}^k\right|^2\right)$.
Further, the time complexity of weighted aggregation based on Eq. \eqref{eq:aggregation} is $\mathcal{O}\left(\left|\mathcal{C}^k\right|\right)$ since it provides $\{\omega_i^k\}$ for each of edge client $i$. 
The complexity of \textbf{Algorithm \ref{alg:algorithm3}} is $\mathcal{O}\left(\left(K-K_p\right)\cdot \left(\left|\mathcal{C}^k\right|^2 + \left|\mathcal{C}^k \right|+ E \cdot n_k \right)\right)$, which can be simplified as $\mathcal{O}\left(K \cdot \left(\left|\mathcal{C}^k\right|^2 + E \cdot n_k\right)\right)$.

In summary, the time complexity of the {\tt FedCMD} framework is governed by the number of edge clients $|\mathcal{C}^k|$, the size of their local datasets $n_k$, the number of epochs $E$, and the number of communication rounds $K$.

\section{Performance Evaluation}
In this section, we introduce the experimental settings, datasets, baselines, and implementation.
We evaluate the performance of our solution by comparing it with nine other methods across accuracy, scalability, ablation experiment, class-wise accuracy analysis, and communication overhead.

\subsection{Experiment Setup and Datasets}\label{section:exp_setup}
We build the federated learning prototype based on an open-source benchmark\footnote{https://github.com/KarhouTam/FL-bench}. All experiments are conducted on a high-performance server with GPU (NVIDIA GeForce RTX 4090 24GB) and CPU (Intel i9-13900K, 24-Core, 3.00GHz), using Pytorch 3.10.13. The code is available at repository\footnote{https://github.com/elegy112138/FedCMD}.

\paragraph{Datasets} Our experiments span ten open benchmark datasets: CIFAR-10, CIFAR-100, CINIC10, EMNIST, FMNIST, MedmNistA, MedmNistC, MNIST, SVHN, and Tiny ImageNet. For further details, please refer to Appendix \ref{appendixA}. Each dataset was divided into 100 subsets (\textit{i.e.}, $n=100$), following the Dirichlet distribution \textit{Dir($\alpha$)} with $\alpha$ set to values within \{0.1, 0.5, 1.0\}, to simulate scenarios with non-IID data. The degree of non-IID is determined by $\alpha$, where a smaller $\alpha$ value indicates more significant data heterogeneity. For example, at $\alpha=0.1$, the non-IID degree is quite pronounced, suggesting that the data held by any specific client is unlikely to cover all classes (\textit{i.e.}, $|\mathcal{Y}_i| \leq |\mathcal{Y}|$), where $\mathcal{Y}_i$ represents the label space of data distributed to the $i$-th client.

\begin{table*}
    \footnotesize 
    \centering
    \caption{Average model accuracy over datasets for \( \alpha = 0.1 \). Bold and underlined indicate the best and second-best respectively.}
    \begin{tabular}{l|c|c|c|c|c|c|c|c|c|c}
        \toprule
        Method & CIFAR10 & CIFAR100 & CINIC10 & EMNIST & FMNIST & MedmNistA & MedmNistC & MNIST & SVHN & T. ImageNet \\
        \midrule
        Local-Only & \underline{85.356} & 36.058 & 78.876 & 92.712 & 95.528 & 91.808 & 90.077 & 96.555 & 89.402 & 21.424 \\
        FedAvg~\cite{mcmahan2017communication} & 24.539 & 13.845 & 29.516 & 75.648 & 77.701 & 67.475 & 67.814 & 94.668 & 68.894 & 9.35 \\
        \hline
        FedBN~\cite{li2021fedbn} & 41.496 & 13.803 & 42.703 & 77.749 & 79.463 & 76.026 & 70.051 & 95.503 & 73.858 & 10.721 \\
        FedAP~\cite{lu2022personalized} & 84.092 & 30.184 & 75.67 & 92.829 & 95.383 & \underline{94.598} & 90.937 & 98.452 & 92.907 & 14.412 \\
        FedBabu~\cite{oh2021fedbabu} & 29.015 & 5.718 & 23.864 & 73.819 & 74.808 & 64.034 & 66.693 & 94.32 & 67.33 & 3.467 \\
        FedFomo~\cite{zhang2020personalized} & 84.081 & 35.754 & 76.455 & 92.411 & 95.341 & 91.351 & 89.712 & 96.561 & 88.517 & 20.513 \\
        FedProx~\cite{li2020federated} & 28.893 & 12.897 & 29.333 & 74.547 & 79.285 & 66.074 & 67.766 & 93.47 & 70.442 & 10.116 \\
        FedRep~\cite{collins2023exploiting} & 85.129 & 36.5 & \textbf{82.914} & \underline{94.379} & 95.501 & 91.691 & 88.662 & 97.852 & 91.531 & 26.066 \\
        pFedSim~\cite{tan2023pfedsim} & 81.712 & \underline{38.452} & 74.771 & \textbf{94.552} & \underline{96.171} & 94.113 & 92.594 & \underline{99.062} & \underline{93.371} & \underline{30.179} \\
        FedDyn~\cite{acar2021federated} & 27.538 & 13.475 & 30.278 & 75.569 & 78.536 & 66.563 & 65.816 & 94.369 & 72.487 & 9.423 \\
        FedPer~\cite{arivazhagan2019federated} & 81.779 & 36.777 & 78.905 & 93.949 & 95.709 & 93.903 & \underline{92.853} & 98.8 & 93.176 & 26.083 \\
        \hline
        FedCMD & \textbf{87.252} & \textbf{48.061} & \underline{80.773} & 94.245 & \textbf{96.569} & \textbf{96.464} & \textbf{94.588} & \textbf{99.441} & \textbf{95.023} & \textbf{31.107} \\
        \bottomrule
    \end{tabular}
    \label{tab:comparision_alpha_0.1}
\end{table*}

\begin{table*}
    \footnotesize 
    \centering
    \caption{Average model accuracy over datasets for \( \alpha = 0.5 \). Bold and underlined indicate the best and second-best respectively.}
    \begin{tabular}{l|c|c|c|c|c|c|c|c|c|c}
        \toprule
        & CIFAR10 & CIFAR100 & CINIC10 & EMNIST & FMNIST & MedmNistA & MedmNistC & MNIST & SVHN & T. ImageNet\\
        \midrule
        Local-Only & 57.366 & 14.37 & 55.863 & 84.462 & 87.024 & 79.494 & 74.989 & 92.377 & 74.408 & 6.646 \\
        FedAvg~\cite{mcmahan2017communication} & 44.085 & 17.393 & 41.550 & 82.636 & 84.491 & 81.696 & 77.488 & 96.908 & 82.54 & 10.207 \\
        \hline
        FedBN~\cite{li2021fedbn} & 50.654 & 16.845 & 46.284 & 82.859 & 84.982 & 85.712 & 81.113 & 97.306 & 83.514 & 11.26 \\
        FedAP~\cite{lu2022personalized} & 62.987 & 18.439 & 56.187 & 86.864 & 89.945 & \underline{90.249} & 87.128 & 98.016 & 87.571 & 6.948 \\
        FedBabu~\cite{oh2021fedbabu} & 44.387 & 8.754 & 39.173 & 82.477 & 84.024 & 81.109 & 78.617 & 96.963 & 81.801 & 3.318 \\
        FedFomo~\cite{zhang2020personalized} & 58.419 & 13.3 & 55.029 & 83.659 & 86.611 & 78.381 & 74.932 & 93.385 & 72.33 & 7.951 \\
        FedProx~\cite{li2020federated} & 45.004 & 16.33 & 41.051 & 82.07 & 83.582 & 80.56 & 77.806 & 96.781 & 82.273 & 10.64 \\
        FedRep~\cite{collins2023exploiting} & \underline{64.459} & 15.594 & \textbf{65.658} & 88.551 & 89.622 & 85.346 & 77.669 & 96.606 & 84.967 & 10.325 \\
        pFedSim~\cite{tan2023pfedsim} & 61.379 & \underline{21.335} & 58.836 & \textbf{90.055} & 89.908 & 89.114 & \underline{87.498} & \underline{98.055} & \underline{88.214} & \textbf{16.725} \\
        FedDyn~\cite{acar2021federated} & 43.818 & 16.861 & 40.815 & 82.711 & 84.11 & 81.278 & 77.674 & 97.021 & 82.038 & 10.085 \\
        FedPer~\cite{arivazhagan2019federated} & 61.908 & 16.908 & 62.104 & 88.032 & \underline{90.034} & 88.544 & 87.367 & 97.67 & 87.592 & 10.615 \\
        \hline
        FedCMD & \textbf{71.331} & \textbf{27.531} & \underline{64.409} & \underline{89.596} & \textbf{92.26} & \textbf{93.982} & \textbf{90.735} & \textbf{98.636} &  \textbf{90.039} & \underline{14.561} \\
        \bottomrule
    \end{tabular}
    \label{tab:comparision_alpha_0.5}
\end{table*}

\begin{table*}
    \footnotesize 
    \centering
    \caption{Average model accuracy over datasets for \( \alpha = 1.0 \). Bold and underlined indicate the best and second-best respectively.}
    \begin{tabular}{l|c|c|c|c|c|c|c|c|c|c}
        \toprule
        Method & CIFAR10 & CIFAR100 & CINIC10 & EMNIST & FMNIST & MedmNistA & MedmNistC & MNIST & SVHN & T. ImageNet \\
        \midrule
        Local-Only & 48.019 & 9.27 & 44.726 & 79.627 & 82.121 & 74.161 & 68.507 & 90.564 & 70.358 & 4.037 \\
        FedAvg~\cite{mcmahan2017communication} & 45.101 & \underline{17.318} & 42.239 & 83.254 & 84.613 & 86.537 & 82.310 & 96.927 & 82.969 & 10.274 \\
        \hline
        FedBN~\cite{li2021fedbn} & 47.935 & 16.756 & 44.147 & 83.012 & 84.841 & 87.967 & 83.808 & 97.218 & 84.108 & 11.462 \\
        FedAP~\cite{lu2022personalized} & 56.092 & 15.853 & 47.755 & 83.921 & 86.848 & \underline{90.233} & \underline{86.504} & \underline{97.59} & 87.33 & 5.847 \\
        FedBabu~\cite{oh2021fedbabu} & 44.735 & 9.025 & 43.988 & 82.806 & 84.436 & 85.716 & 80.235 & 96.776 & 82.582 & 3.63 \\
        FedFomo~\cite{zhang2020personalized} & 46.625 & 8.84 & 44.022 & 78.865 & 82.282 & 73.107 & 73.981 & 93.311 & 68.423 & 5.376 \\
        FedProx~\cite{li2020federated} & 44.853 & 16.665 & 43.028 & 82.443 & 84.515 & 86.496 & 82.333 & 96.88 & 82.792 & 10.637 \\
        FedRep~\cite{collins2023exploiting} & 55.345 & 10.167 & \textbf{57.861} & 84.444 & 86.088 & 83.942 & 74.709 & 95.976 & 84.163 & 6.77 \\
        pFedSim~\cite{tan2023pfedsim} & 56.454 & 17.055 & 52.834 & \textbf{87.203} & \underline{88.276} & 88.296 & 86.434 & 97.587 & \underline{87.828} & \textbf{12.905} \\
        FedDyn~\cite{acar2021federated} & 45.964 & 17.016 & 42.826 & 83.217 & 84.754 & 86.851 & 82.584 & 97.042 & 83.229 & 10.271 \\
        FedPer~\cite{arivazhagan2019federated} & \underline{56.753} & 12.477 & 55.777 & 84.258 & 88.018 & 87.846 & 85.892 & 97.198 & 87.066 & 7.91 \\
        \hline
        FedCMD & \textbf{64.508} & \textbf{22.178} & \underline{57.468} & \underline{86.454} & \textbf{89.837} & \textbf{93.312} & \textbf{89.448} & \textbf{98.271} & \textbf{89.464} & \underline{12.537} \\
        \bottomrule
    \end{tabular}
    \vspace{-2em}
    \label{tab:comparision_alpha_1}
\end{table*}

\paragraph{Baselines} We compare {\tt FedCMD} with nine state-of-the-art federated learning solutions. To create a baseline for evaluating the generalization and personalization performance, respectively, \textcolor{black}{we implement {\tt FedAvg}~\cite{mcmahan2017communication} and Local-training-only in our experiments.} Local-training-only means that the clients only update parameters with their local data without the global parameter aggregation. In addition, we achieved baseline comparisons of the following 9 algorithms, and see Appendix~\ref{appendixA} for detailed hyper-parameter settings: 
\begin {itemize}[\itemindent=0pt]
\item {\tt FedProx}~\cite{li2020federated} adds a proximal term to the loss function to prevent the edge client model from drifting away from the global model.
\item {\tt FedRep}~\cite{collins2023exploiting} treats the classifier as the personalized layer, and trains the classifier locally and the feature extractor sequentially.
\item {\tt FedBabu}~\cite{oh2021fedbabu} initializes randomly the classification header and updates the body only during training.
\item {\tt pFedSim}~\cite{tan2023pfedsim} uses classifier distance as the weights for global aggregation and based on model decoupling.
\item {\tt FedPer}~\cite{arivazhagan2019federated} employs the last-layer-as-head decoupling by splitting the network into the base and personalized layer to achieve personalization.
\item {\tt FedBN}~\cite{li2021fedbn} preserves batch normalization layers in the edge client model locally to stabilize training and the rest uploads to aggregate.
\item {\tt FedAP}~\cite{lu2022personalized} measures client-wised similarity using Wasserstein distance based on statistics from the batch normalization layer to optimize model aggregation.
\item {\tt FedDyn}~\cite{acar2021federated} aligns the solutions of each edge client and server utilizing a dynamic regularizer from a communication perspective.
\item {\tt FedFomo}~\cite{zhang2020personalized} calculates how much one client can benefit from the other, deriving the optimally weighted model combination for each client.
\end {itemize}

\paragraph{Implementation} We implemented LeNet5~\cite{lecun1998gradient} for all methods to conduct performance evaluation on all datasets. According to the setup in~\cite{tan2023pfedsim}, we evenly split the dataset of each edge client into the training and test sets with no intersection, \textit{i.e.}, $|\mathcal{D}_i^{train}|=|\mathcal{D}_i^{test}|$. On the edge client side, we adopt SGD as the local optimizer. The local learning rate is set to 0.01 for all experiments. All experiments shared the same \textcolor{black}{join ratio $\gamma$ equals to 0.1}, communication round $K=200$, local epoch $E=5$, and batch size equals to 32. 
In Appendix~\ref{appendixA}, we provide the list of the model architectures and the full hyperparameter configurations employed in the methods involved.

\subsection{Comparison of Classification Accuracy}
We compare the average model accuracy performance of our solution with other baselines in Table~\ref{tab:comparision_alpha_0.1}--\ref{tab:comparision_alpha_1}. 
The results displayed in Table \ref{tab:comparision_alpha_0.1} compare the average accuracy with \(\alpha = 0.1\), where {\tt FedCMD} exhibits outstanding performance across most datasets. Notably,  {\tt FedCMD} leads with significant margins in complex datasets like CIFAR100 and Tiny ImageNet, with remarkable accuracy improvements of 9.609\% and 0.928\% respectively over the second-best results. 
Compared to established methods like {\tt FedBN}, {\tt FedBabu}, {\tt FedProx} and {\tt FedDyn},  {\tt FedCMD} shows a marked improvement. For example, in the CIFAR10 dataset,  our solution achieves an accuracy of 87.252\%, which is 1.896\% higher than the second-best, Local-Only. Similarly, in MedmNistC, our solution attains 94.588\%, outperforming the nearest competitor, {\tt FedPer}, by 1.735\%.
The underlined results indicate the second-best performing methods, which offer strong competition in datasets like CINIC10. For instance, {\tt FedRep} demonstrates robust performance in CINIC10 with an accuracy of 82.914\%. These results suggest that while other methods have merits in specific conditions,  {\tt FedCMD} consistently leads in adapting to non-IID data in most federated learning environments.

In Table \ref{tab:comparision_alpha_0.5}, we see average accuracy with \(\alpha = 0.5\), indicating less data heterogeneity compared to \(\alpha = 0.1\). 
In this context, the performance of our {\tt FedCMD} method, while still leading in many cases. 
Despite this, {\tt FedCMD} remains the top performer in several datasets, notably achieving the highest accuracy in CIFAR10 at 71.331\%, which is 6.872\% higher than the second-best, {\tt FedRep}. 
In CIFAR100, {\tt FedCMD} leads with 27.531\%, outperforming {\tt pFedSim}, the second-best, by a significant margin of 6.196\%. This trend continues in Tiny ImageNet, where {\tt FedCMD}'s accuracy of 14.561\% is second only to {\tt pFedSim}'s 16.725\%. 
Compared with other methods, our solution's performance in datasets like FMNIST and MedmNistA is notably superior, with accuracy of 92.260\% and 93.982\% respectively, demonstrating its effectiveness even in less heterogeneous settings. 
Despite the reduced heterogeneity, our solution continues to demonstrate robust performance across ten datasets with \(\alpha = 0.5\). 

Table \ref{tab:comparision_alpha_1} shows the average model accuracy with \(\alpha = 1.0\), indicating a more uniform data distribution than \(\alpha = 0.1\) and \(\alpha = 0.5\). 
Despite this, {\tt FedCMD} continues to lead in several datasets. 
Notably, it achieves the highest accuracy in CIFAR10 at 64.508\%, surpassing the second-best {\tt FedPer} by 7.755\%. In CIFAR100, {\tt FedCMD} tops the list with an accuracy of 22.178\%, showing a noticeable improvement over {\tt FedAvg}. In datasets like FMNIST and MedmNistA, our solution outperforms others with the accuracy of 89.837\% and 93.312\%, respectively, underlining its effectiveness across varied settings. 
{\tt FedCMD} also demonstrates strong performance in Tiny ImageNet, achieving 12.537\%, which, although not the highest, is close to the top performer, {\tt pFedSim}. Compared to {\tt FedDyn}, our solution shows a considerable advantage in datasets like CIFAR10 and CIFAR100, but a narrower lead in more uniform datasets like EMNIST and MNIST. 
Overall, while {\tt FedCMD}'s performance advantage diminishes in less heterogeneous scenarios (\(\alpha = 1.0\)), it still shows notable effectiveness and robustness across a range of datasets. 

\begin{figure*}[htbp]
    \centering
    \begin{minipage}[b]{1\linewidth}
        \centering
        \includegraphics[width=\linewidth]{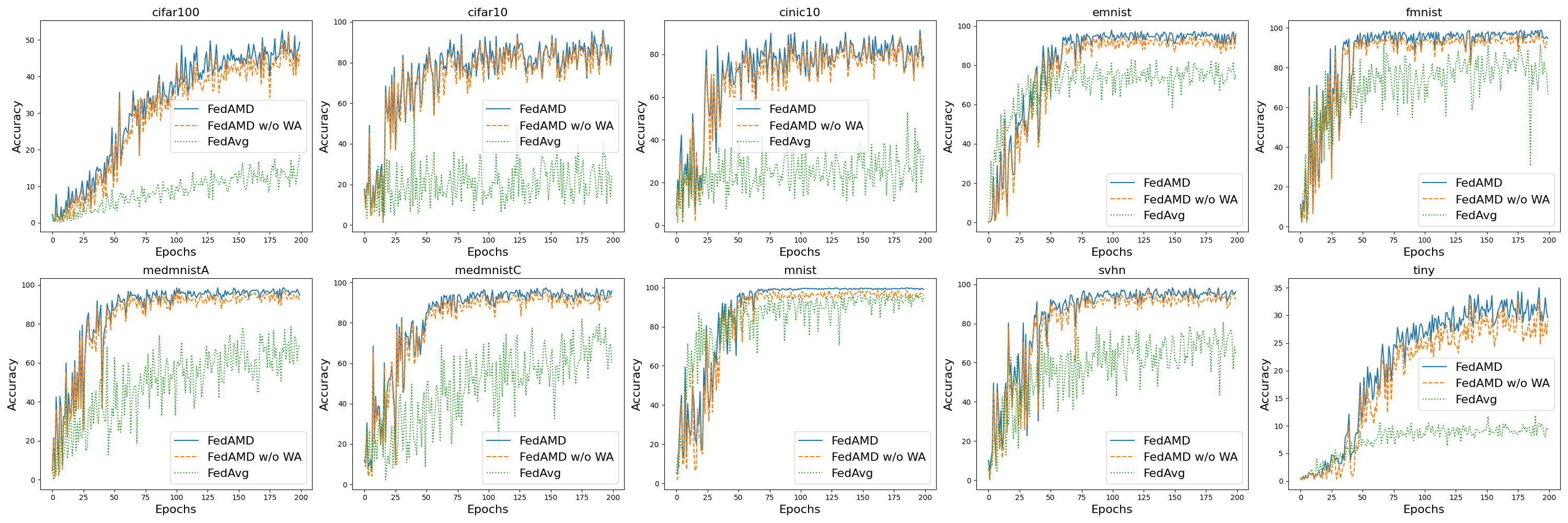}
    \end{minipage}
    \vspace{-1.5em}
    \caption{Accuracy comparison of {\tt FedCMD}, {\tt FedCMD} without weighted aggregation, and original {\tt FedAvg} across ten datasets.}
    \vspace{-1.5em}
    \label{fig:ablation}
\end{figure*}

\begin{figure}[htbp]
    \centering
    \begin{minipage}[b]{1\linewidth}
        \centering        
        \includegraphics[width=\linewidth]{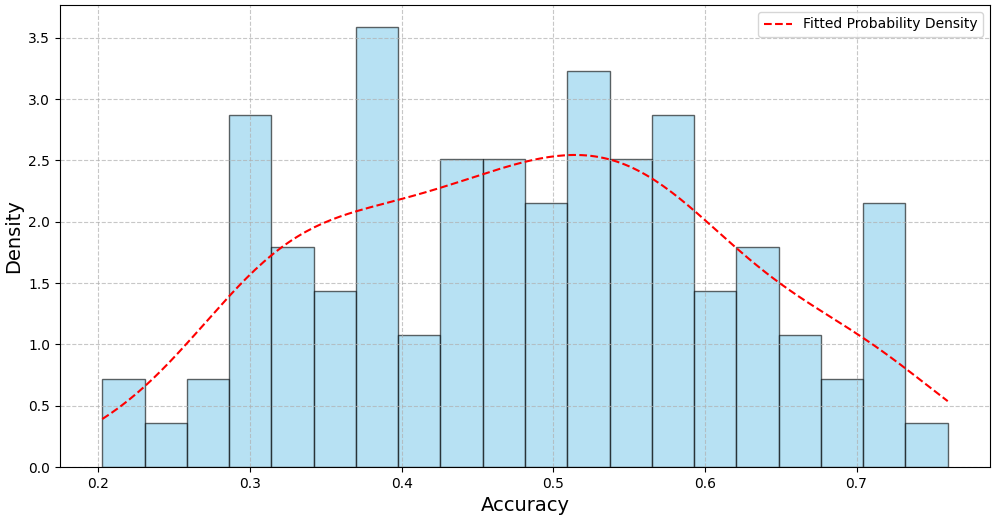}
    \end{minipage}
    \vspace{-1.5em}
    \caption{Probability Density Function of Accuracy on CIFAR100.}
    \vspace{-2em}
    \label{fig:PDF}
\end{figure}

\subsection{Class-wise Accuracy Analysis}
Fig. \ref{fig:PDF} shows the probability density function of class-wise accuracy on CIFAR100 with {\tt FedCMD}.
CIFAR100 consists of 100 distinct classes, and this analysis involves calculating the accuracy for each individual class. The probability density histogram in the figure is based on these 100 accuracy values, along with a fitted probability density curve to visualize the distribution pattern.
From the histogram, it's evident that the accuracy distribution for the 100 classes lies within the range of 20\% to 75\%. A significant concentration of class accuracies is observed around 50\%, indicating a median level of recognition capability for most classes. However, the graph also reveals the presence of a few classes with notably lower accuracy, below 30\%, as well as some with higher accuracy, exceeding 70\%.
This distribution suggests a varied level of difficulty in accurately recognizing different classes within CIFAR100. While a majority of the classes are recognized with moderate accuracy, a subset of classes poses greater challenges or are recognized with high accuracy.

\subsection{Ablation Experiment}
As illustrated in Fig. \ref{fig:ablation}, we present the results of our ablation experiments. These experiments compare three designs: {\tt FedCMD} (which includes both weighted aggregation and model decoupling), {\tt FedCMD} without weighted aggregation (employing only model decoupling), and the original {\tt FedAvg} method that doesn't utilize either weighted aggregation or model decoupling.
The comparison between {\tt FedCMD} and {\tt FedCMD} without weighted aggregation reveals that the inclusion of weighted aggregation in {\tt FedCMD} contributes to a modest performance enhancement, with an average improvement of approximately 2\% across all ten scenarios. 
However, the most significant performance leap is observed when comparing {\tt FedCMD} without weighted aggregation to the original {\tt FedAvg}. The absence of model decoupling and weighted aggregation in {\tt FedAvg} leads to substantially lower performance. For instance, in the CIFAR10 dataset, the implementation of model decoupling in {\tt FedCMD} without weighted aggregation results in nearly a 60\% increase in accuracy. Even in the MNIST dataset, there's at least a 3\% improvement in performance. These results show the importance of model decoupling in the performance enhancement.

\begin{figure*}[htbp]
    \centering
    \begin{minipage}[b]{1\linewidth}
        \centering        
        \includegraphics[width=\linewidth]{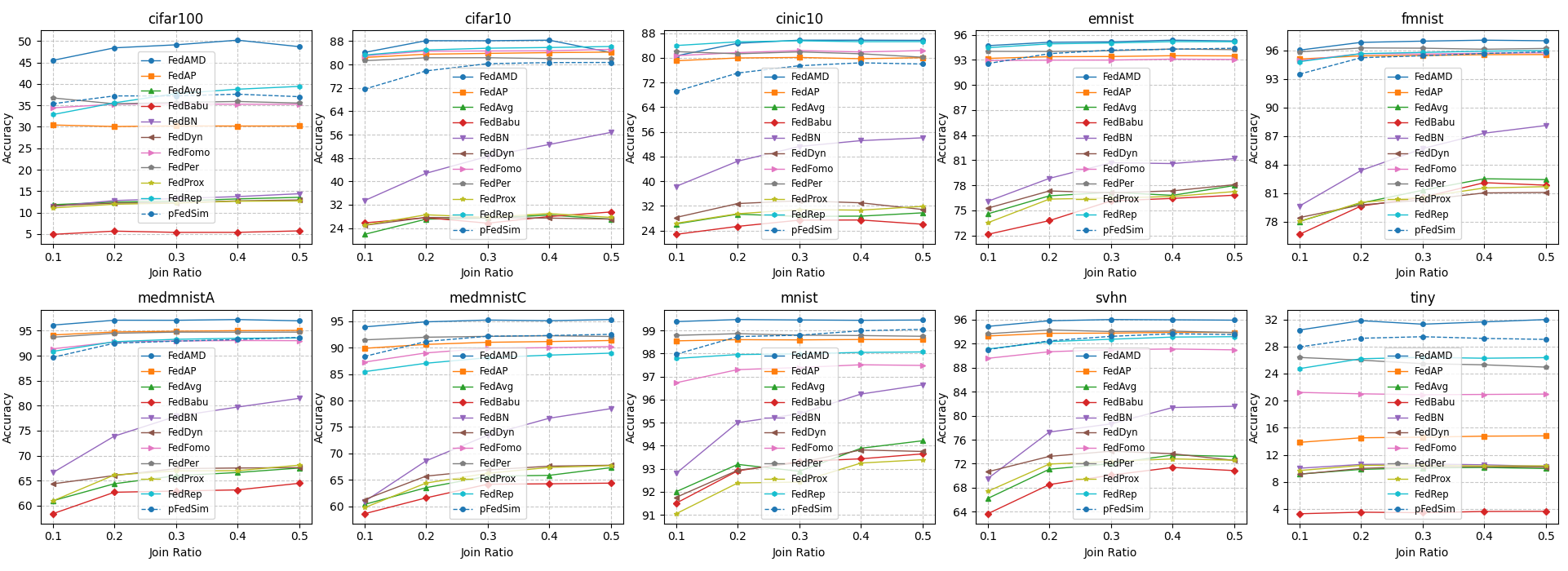}
    \end{minipage}
    \vspace{-1.5em}
    \caption{Accuracy comparison for {\tt FedCMD} and ten solutions (including {\tt FedAvg}) with different join ratios across ten datasets.}
    \vspace{-1.5em}
    \label{fig:scalability}
\end{figure*}

\subsection{Comparison of Scalability}

As shown in Fig. \ref{fig:scalability}, the performance of various federated learning, including our {\tt FedCMD}, was evaluated under different client join ratios \{0.1, 0.2, 0.3, 0.4, 0.5\}. 
Our {\tt FedCMD} method consistently showed performance improvement with increasing join ratios in most of the datasets, such as in CIFAR-10, starting from 84.114\% at a ratio of 0.1 and peaking at 88.331\% at a ratio of 0.4. This trend indicates that {\tt FedCMD} effectively leverages increased client participation, enhancing its performance.
Comparatively, other methods like {\tt FedBN}, {\tt pFedSim}, and {\tt FedAP} also exhibited performance boosts with higher join ratios, but none matched the level of improvement seen with {\tt FedCMD}. For instance, in CIFAR-10, {\tt FedBN}'s accuracy improved from 33.388\% to 52.618\%, and {\tt pFedSim} increased from 71.559\% to 80.653\%. {\tt FedAP} showed a consistent rise from 82.404\% to 84.065\%.
Methods like {\tt FedBabu}, {\tt FedProx}, and {\tt FedDyn}, however, demonstrated relatively minor improvements or fluctuations across different join ratios. {\tt FedBabu}'s performance fluctuated around the mid-20\% range, while {\tt FedProx} and {\tt FedDyn} hovered around the high-20\% to low-30\% range.
{\tt FedFomo}, {\tt FedRep}, and {\tt FedPer} also displayed improvements with increasing join ratios, but their increments were not as substantial as {\tt FedCMD}. For example, in CIFAR-10, {\tt FedFomo} and {\tt FedRep} started at 82.942\% and 83.257\% at a ratio of 0.1, respectively, and increased to around 84.680\% and 85.822\% at a ratio of 0.4.
Interestingly, {\tt FedBabu}'s performance improves with an increase in the joining ratio but declines at the highest ratio of 0.5, suggesting a threshold beyond which more client participation does not benefit.
Overall, the data demonstrates that our {\tt FedCMD} method not only starts off with higher accuracy at lower join ratios but also scales more effectively with increased client participation compared to other methods.

\subsection{Disscussion of Layer Selection Phase}
Our study on the impact of communication rounds for the personalized layer selection phase in {\tt FedCMD} reveals the relationship between the number of rounds and model accuracy. The data from Table \ref{tab:param_rho} shows that while most datasets benefit from an increased number of rounds, there are exceptions where fewer rounds also yield better results.
In most cases, a trend of increasing accuracy with more communication rounds is evident. For instance, CIFAR10 peaks at 87.51\% with 80 rounds, FMNIST reaches 96.775\% with 100 rounds, and Tiny ImageNet consistently improves, achieving its highest accuracy of 31.602\% with 100 rounds. This trend suggests that more communication rounds generally lead to more accurate personalized layers, enhancing overall model performance.
However, some datasets such as EMNIST peak at 94.912\% with just 20 rounds, indicating that fewer rounds can be more effective in certain scenarios. Similarly, CIFAR100 and MedmNistC show optimal performances at 40 rounds with an accuracy of 47.954\% and 94.745\%, respectively. These exceptions highlight that the relationship between communication rounds and accuracy is not always linear and can vary depending on the dataset's characteristics.
Overall, while the general trend suggests that more communication rounds in {\tt FedCMD} lead to better accuracy, the optimal number of rounds is highly dataset-dependent.

\begin{table}
    \footnotesize
    \centering
    \caption{Accuracy with different rounds of layer selection phase across ten datasets.}
    \begin{tabular}{lccccc}
        \toprule
        Number of rounds & 20 & 40 & 60 & 80 & 100 \\
        \midrule
        CIFAR10 & 84.076 & 87.427 & 87.422 & \textbf{87.51} & 87.117 \\
        CIFAR100 & 42.224 & \textbf{47.954} & 47.552 & 47.374 & 47.527 \\
        CINIC10 & 78.426 & \textbf{80.619} & 79.959 & 79.823 & 79.573 \\
        EMNIST & \textbf{94.912} & 94.221 & 94.225 & 94.165 & 94.295 \\
        FMNIST & 94.736 & 96.516 & 96.705 & 96.508 & \textbf{96.775} \\
        MedmNistA & 95.204 & 96.362 & \textbf{96.517} & 96.418 & 96.444 \\
        MedmNistC & 94.364 & \textbf{94.745} & 94.49 & 94.485 & 94.564 \\
        MNIST & 99.266 & 99.289 & \textbf{99.313} & 99.281 & 99.259 \\
        SVHN & 94.302 & \textbf{95.178} & 95.047 & 94.961 & 94.925 \\
        Tiny ImageNet & 28.877 & 31.17 & 31.498 & 31.494 & \textbf{31.602} \\
        \bottomrule
    \end{tabular}
    \vspace{-2em}
    \label{tab:param_rho}
\end{table}

\subsection{Communication Overhead}

\begin{figure}[htbp]
    \centering
    \begin{minipage}[b]{1\linewidth}
        \centering        
        \includegraphics[width=\linewidth]{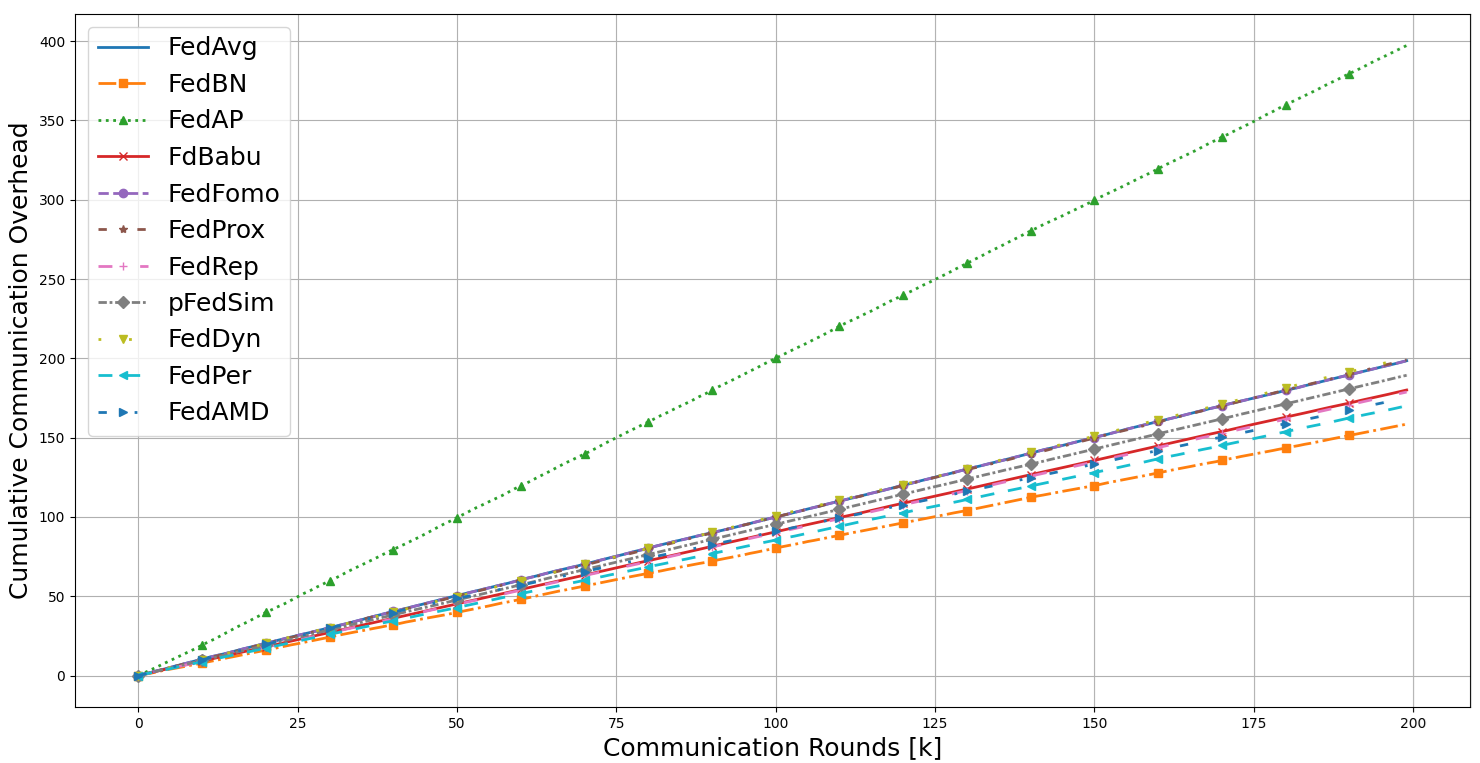}
    \end{minipage}
    \vspace{-1.5em}
        \caption{Comparison of Cumulative Communication Overhead.}
    \vspace{-1.5em}
    \label{fig:communication}
\end{figure}

The experimental results depicted in Fig. \ref{fig:communication} present a comparison of the cumulative communication overhead across ten different algorithms against the benchmark {\tt FedAvg}.
The analysis reveals that most methods, except for {\tt FedAP}, have comparable or even lower communication overhead compared to {\tt FedAvg}. Notably, {\tt FedBN} exhibits the lowest overhead among all. This efficiency is attributed to its approach of updating multiple layers only at the edge, eliminating the need for uploading a substantial amount of parameters to the cloud for aggregation. Similarly, {\tt FedPer} also shows commendable performance due to its strategy for selecting multiple layers as personalizing.
{\tt FedCMD} only ranks behind {\tt FedBN} and {\tt FedPer}.  Since the last layer typically has fewer parameters than other fully connected layers, methods fixing the last layer for personalization inherently have higher communication costs than {\tt FedCMD}. In summary, our {\tt FedCMD} method demonstrates effective communication overhead management.

\section{Conclusions}
In this paper, we present a novel personalized layer selection in federated learning through our novel {\tt FedCMD} framework. This approach challenges the conventional rigidity of existing layer-selection methods by introducing a distribution distance metric, named feature distribution shift, based on the Wasserstein distance. This metric assesses the alignment between each layer's feature distribution and the heterogeneous data distribution of client models.
To our knowledge, our work marks a significant first step in this area, offering a unique perspective on contrastive layer selection. 
The extensive experimental testing across ten datasets not only validates the effectiveness of our proposed metric but also demonstrates the superior performance of the {\tt FedCMD} framework compared to nine other SOTA solutions. 
Moreover, {\tt FedCMD} two-phase structure, encompassing personalized layer selection and heterogeneous federated learning, along with the introduction of practical algorithms for contrastive layer selection and weighted global aggregation. 
Our comprehensive exploration and analysis in this paper pave the way for future advancements in contrastive and heterogeneous federated learning.

\begin{IEEEbiography}[{\includegraphics[width=0.9in,height=1.2in,clip,keepaspectratio]{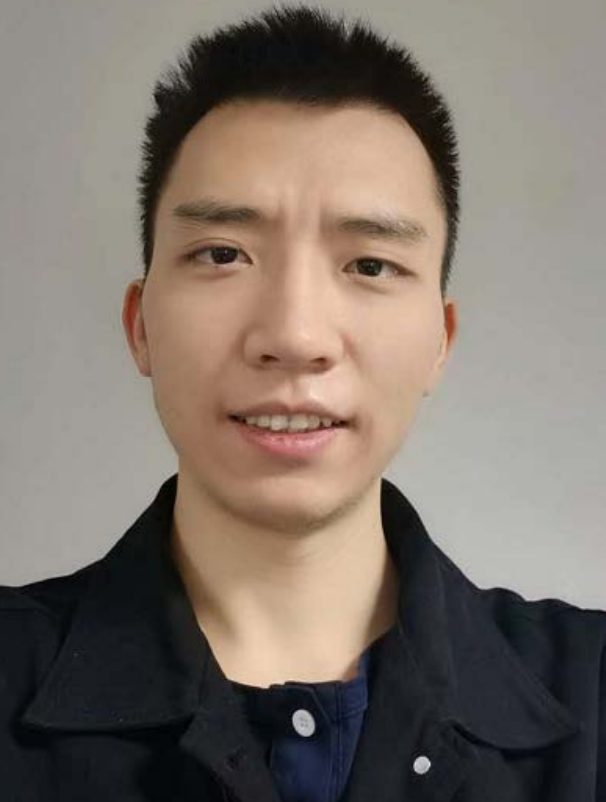}}]{Xingyan Chen}
	received the Ph. D degree in computer technology from Beijing University of Posts and Telecommunications (BUPT), in 2021. 
	He is currently a lecturer with the School of Economic Information Engineering, Southwestern University of Finance and Economics, Chengdu.
	He has published papers in well-archived international journals and proceedings, such as the \textsc{IEEE Transactions on Mobile Computing}, \textsc{IEEE Transactions on Circuits and Systems for Video Technology}, \textsc{IEEE Transactions on Industrial Informatics},  and \textsc{IEEE INFOCOM} etc. 
	His research interests include Multimedia Communications, Multi-agent Reinforcement Learning and Stochastic Optimization.
\end{IEEEbiography}

\vspace{-3em}

\begin{IEEEbiography}
[{\includegraphics[width=0.9in,height=1.2in,clip,keepaspectratio]{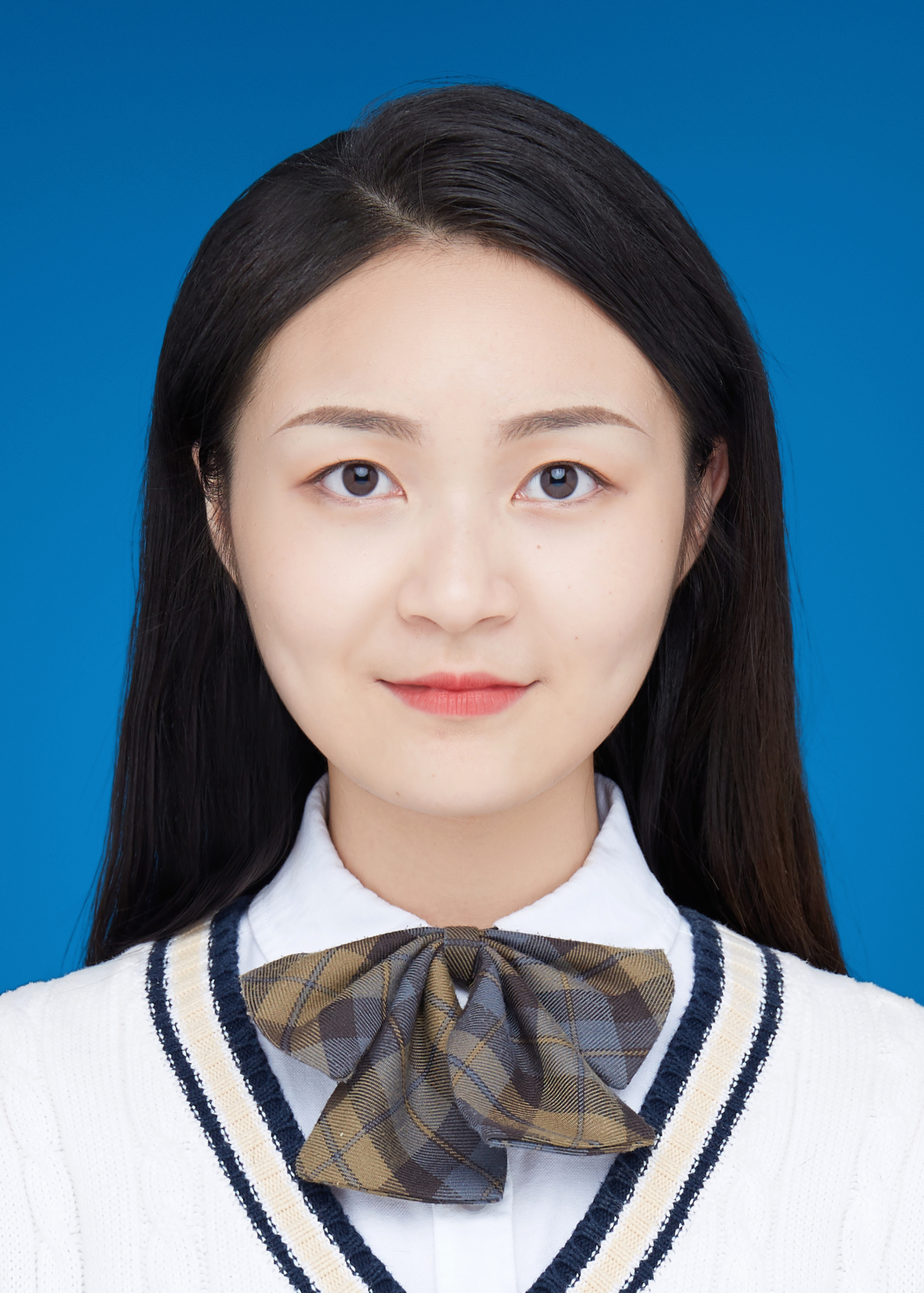}}]{Tian Du} received her B.S. degree in Management from Huaqiao University in 2021, and currently pursuing a Ph.D. in Management specializing in Big Data Management at Southwestern University of Finance and Economics. Her research interests encompass Federated Learning, Reinforcement Learning and Time Series Clustering. She has published in relevant SCI journals. Passionate about exploring cutting-edge data management methodologies and technologies, she is dedicated to addressing intricate challenges related to Big Data, thereby contributing to the advancement of the field of Data Science.

\end{IEEEbiography}

\vspace{-3em}

\begin{IEEEbiography}[{\includegraphics[width=0.9in,height=1.2in,clip,keepaspectratio]{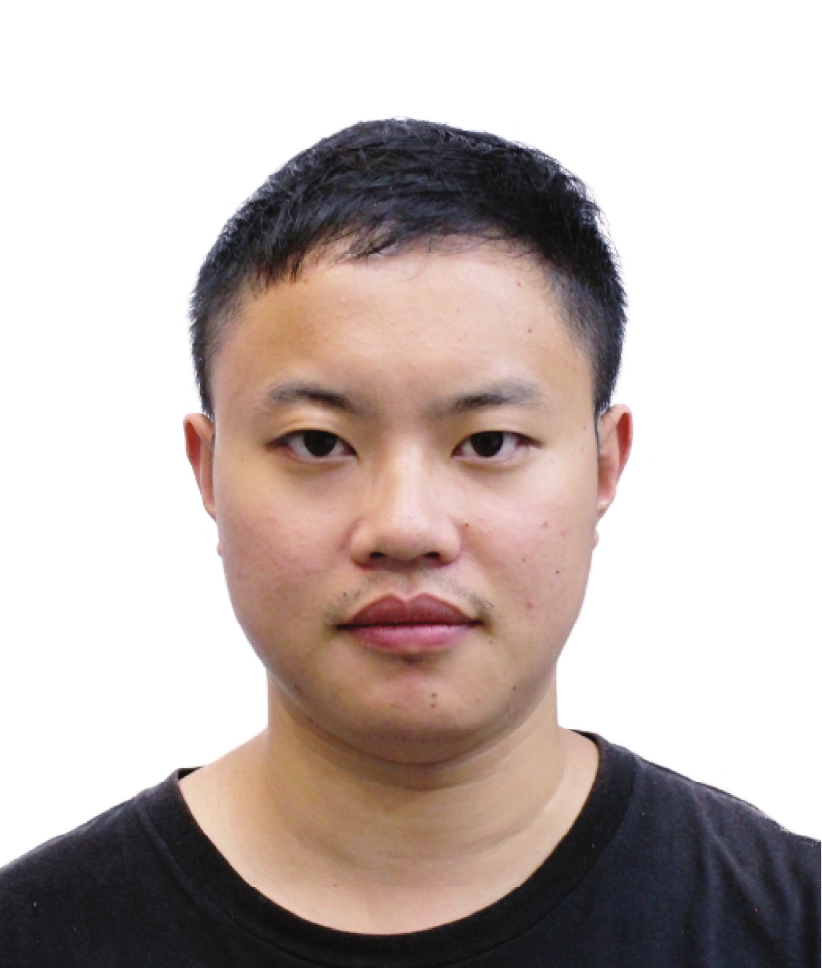}}]{Mu Wang}
	received his M.S. and Ph.D. degrees in computer technology from Beijing University of Posts and Telecommunications (BUPT) in 2015 and 2020. He was a Joint Ph.D. student at the School of Electrical, Computer, and Energy Engineering (ECEE), Arizona State University. He is currently a postdoctoral research associate with the Department of Computer Science and Technology \& BNRist, Tsinghua University. His research interests include information-centric networking, wireless communications, and multimedia delivery in wireless networks.
\end{IEEEbiography}

\vspace{-3em}

\begin{IEEEbiography}[{\includegraphics[width=0.9in,height=1.2in,clip,keepaspectratio]{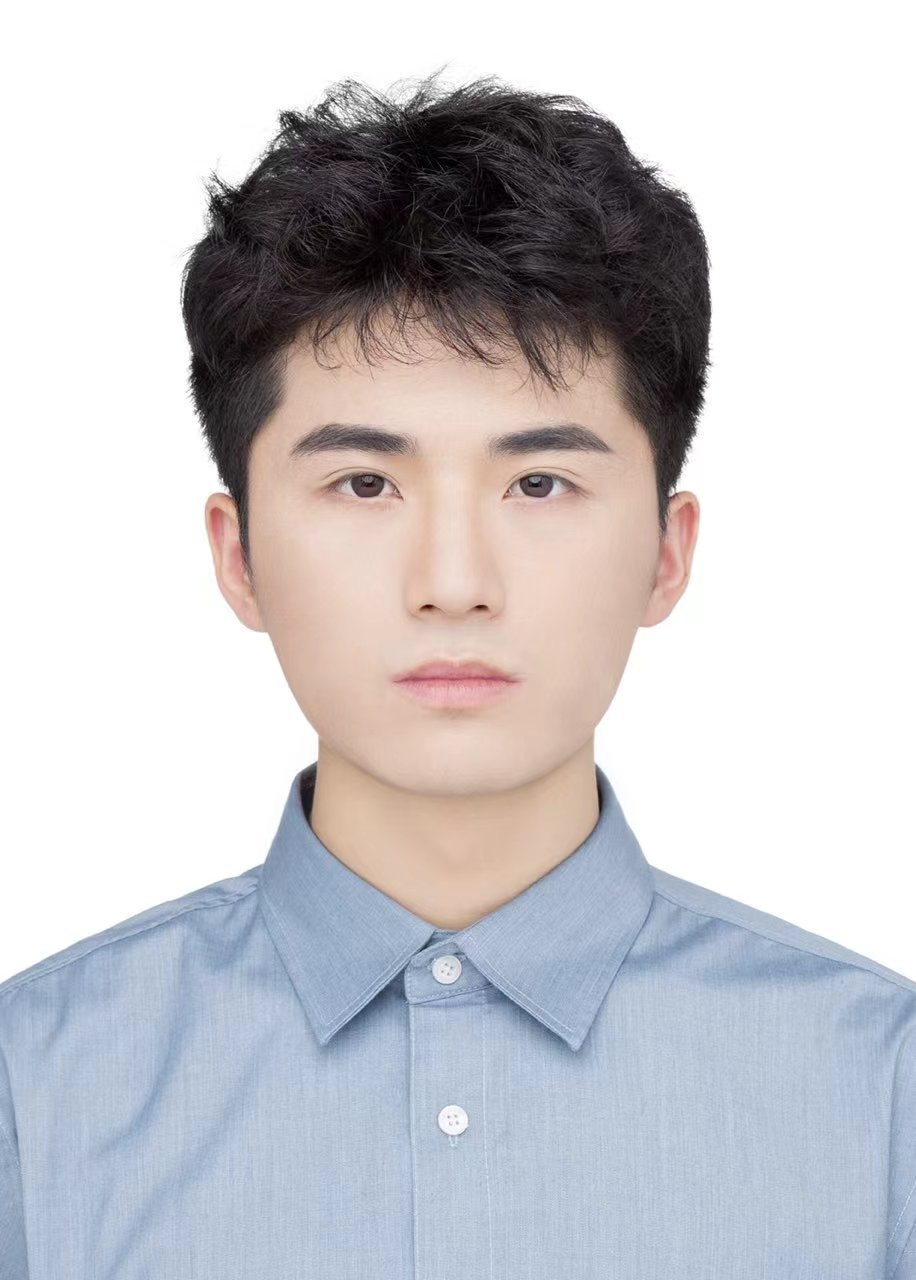}}]{Tiancheng Gu} received his B.S. degree in Engineering from Chongqing University of Posts and Telecommunications in 2023, is currently pursuing a M.S. degree in Computer Science and Technology at Southwestern University of Finance and Economics. His research primarily focuses on large-scale model inference engineering, exploring data analysis and prediction, with a particular emphasis on federated learning and data privacy protection. His career objective is to optimize large-scale model inference techniques based on inference engineering, enhancing data processing efficiency, and contributing to the field of computer science.

\end{IEEEbiography}

\vspace{-3em}

\begin{IEEEbiography}[{\includegraphics[width=1in,height=1.25in,clip,keepaspectratio]{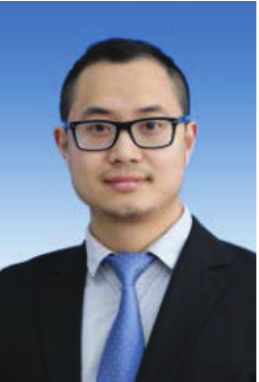}}]{Yu Zhao}
received the B.S. degree from Southwest Jiaotong University in 2006, and the M.S. and Ph.D. degrees from the Beijing University of Posts and Telecommunications in 2011 and 2017, respectively. He is a Professor at the Southwestern University of Finance and Economics. 
His current research interests include natural language processing, large language models, graph learning, and fintech.
He has authored more than 30 papers including IEEE TKDE, IEEE TNNLS, IEEE TMC, IEEE TMM, ACL, ICME, etc.
\end{IEEEbiography}

\vspace{-3em}

\begin{IEEEbiography}[{\includegraphics[width=1in,height=1.25in,clip,keepaspectratio]{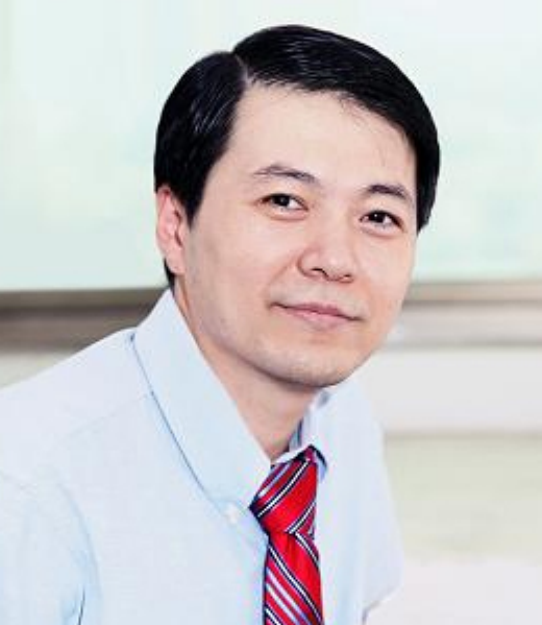}}]{Gang Kou} is a Distinguished Professor of Chang Jiang Scholars Program in Southwestern University of Finance and Economics, managing editor of International Journal of Information Technology \& Decision Making (SCI) and managing editor-in-chief of Financial Innovation (SSCI). He is also editors for other journals, such as: Decision Support Systems, and European Journal of Operational Research. Previously, he was a professor of School of Management and Economics, University of Electronic Science and Technology of China, and a research scientist in Thomson Co., R \& D. He received his Ph.D. in Information Technology from the College of Information Science \& Technology, Univ. of Nebraska at Omaha; Master degree in Dept of Computer Science, Univ. of Nebraska at Omaha; and B.S. degree in Department of Physics, Tsinghua University, China. He has published more than 100 papers in various peer-reviewed journals. Gang Kou’s h-index is 57 and his papers have been cited for more than 10000 times. He is listed as the Highly Cited Researcher by Clarivate Analytics (Web of Science).
\end{IEEEbiography}

\vspace{-3em}

\begin{IEEEbiography}[{\includegraphics[width=1in,height=1.2in,clip,keepaspectratio]{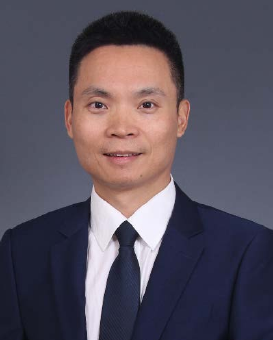}}]{Changqiao Xu} [SM'15] received the Ph.D. degree from the Institute of Software, Chinese Academy of Sciences (ISCAS) in Jan. 2009. He was an Assistant Research Fellow and R\&D Project Manager in ISCAS from 2002 to 2007. He was a researcher at Athlone Institute of Technology and Joint PhD at Dublin City University, Ireland during 2007-2009. He joined Beijing University of Posts and Telecommunications (BUPT), Beijing, China, in Dec. 2009. Currently, he is a full Professor with the State Key Laboratory of Networking and Switching Technology, and Director of the Network Architecture Research Center at BUPT. His research interests include Future Internet Technology, Mobile Networking, Multimedia Communications, and Network Security. He has published over 160 technical papers in prestigious international journals and conferences, including IEEE Comm. Surveys \& Tutorials, IEEE Wireless Comm., IEEE Comm. Magazine, IEEE/ACM ToN, IEEE TMC etc. He has served a number of international conferences and workshops as a Co-Chair and Technical Program Committee member. He is currently serving as the Editorin-Chief of Transactions on Emerging Telecommunications Technologies (Wiley). He is Senior member of IEEE.
\end{IEEEbiography}

\vspace{-3em}

\begin{IEEEbiography}[{\includegraphics[width=1in,height=1.2in,clip,keepaspectratio]{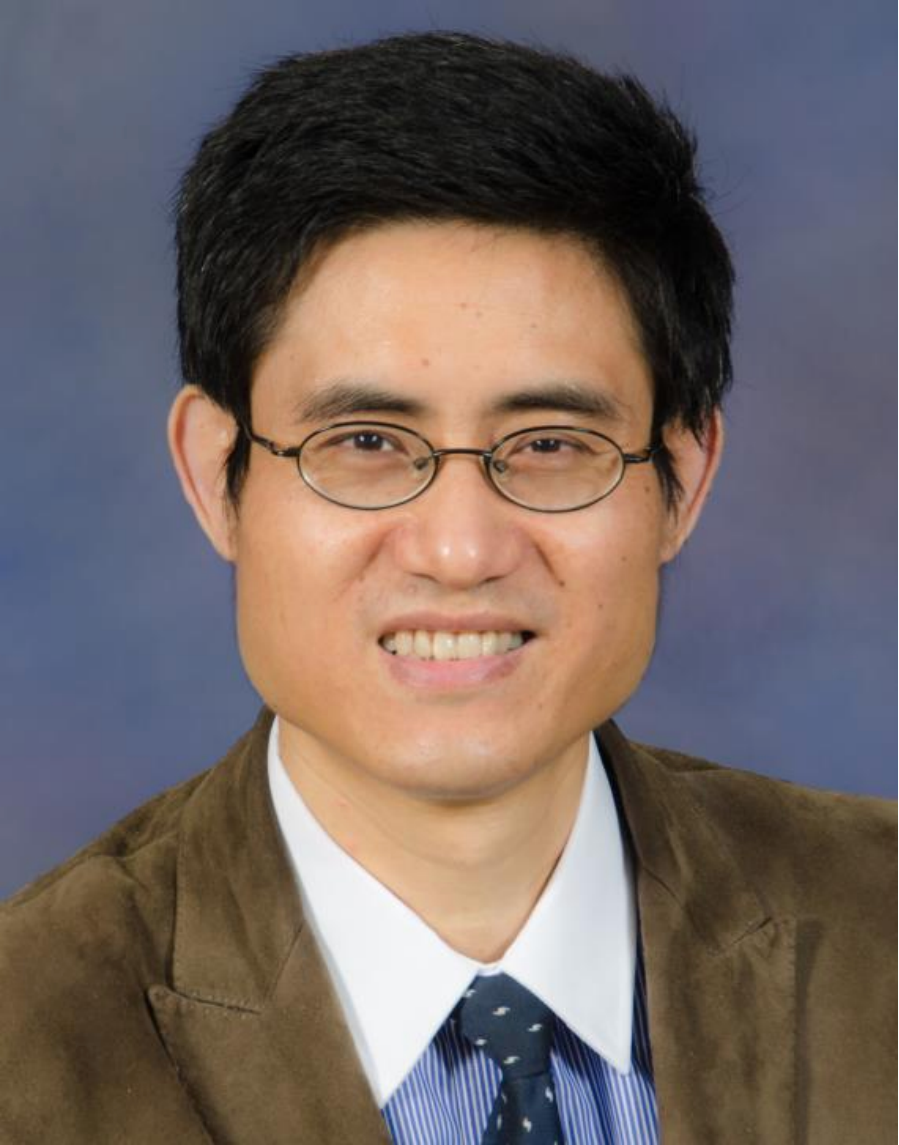}}]{Dapeng Oliver Wu} [S’98-M’04-SM’06-F’13] received a B.E. degree in electrical engineering from Huazhong University of Science and Technology,Wuhan, China in 1990, an M.E. degree in electrical engineering from Beijing University of Posts and Telecommunications, Beijing, China in 1997, and a Ph.D. degree in electrical and computer engineering from Carnegie Mellon University, Pittsburgh, PA in 2003. He is currently Yeung Kin Man Chair Professor of Network Science, and Chair Professor of Data Engineering at the Department of Computer Science, City University of Hong Kong. Previously, he was on the faculty of University of Florida, Gainesville, FL, USA and was the director of NSF Center for Big Learning, USA. His research interests are in the areas of networking, communications, signal processing, computer vision, machine learning, smart grid, and information and network security. He has served as Editor in Chief of IEEE Transactions on Network Science and Engineering, Editor-at-Large for IEEE Open Journal of the Communications Society, founding Editor-in-Chief of Journal of Advances in Multimedia, and Associate Editor for IEEE Transactions on Cloud Computing, IEEE Transactions on Communications, IEEE Transactions on Signal and Information Processing over Networks, IEEE Signal Processing Magazine, IEEE Transactions on Circuits and Systems for Video Technology, IEEE Transactions on Wireless Communications and IEEE Transactions on Vehicular Technology. He is an IEEE Fellow.
\end{IEEEbiography}

\newpage
\clearpage

\appendix
\subsection{Experiment Details} \label{appendixA}
\subsubsection{Computing Configuration}
We implement our experiments using PyTorch 2.1.0+cu121 and run all experiments on Microsoft Windows 11 Professional Edition. Our system configuration includes a 13th Gen Intel(R) Core(TM) i9-13900K CPU @ 3.00GHz with 24 cores and 32 logical processors, combined with an NVIDIA GeForce RTX 4090 GPU. The system is built on a Micro-Star International Co., Ltd. MS-7D25 model (PRO Z690-A WIFI DDR4) with 64.0 GB of RAM.

\subsubsection{Datasets Details}
\begin{table*}
    \centering
    \begin{tabular}{cccccc}
        \toprule
        Dataset & Samples & Classes & Description & Resolution & Year \\
        \midrule
        CIFAR10/100  & 60,000  & 10/100 & Images in 10/100 classes including airplanes, cars, birds, etc. & $32\times32$ & 2009 \\
        CINIC10  & 270,000  & 10 & Images in 10 classes including airplanes, cars, birds, etc. & $32\times32$ & 2018 \\
        EMNIST  & 805,263  & 62 & Extended MNIST with letters and digits & $28\times28$ & 2017 \\
        FMNIST & 70,000 & 10 & Fashion item images & $28\times28$ & 2017 \\
        MNIST & 70000 & 10 & Handwritten digits & $28\times28$ & 1998 \\
        MedmNistA/C & 58,850/23,600 & 11 & Biomedical images on abdominal CT. & $28\times28$ & 2019 \\
        SVHN & 600,000 & 10 & Street view house numbers & $32\times32$ & 2011 \\
        Tiny ImageNet & 110,000 & 200 & Tiny ImageNet dataset & $64\times64$ & 2015\\
        \bottomrule
    \end{tabular}
    \caption{Statistical information of used datasets on clients.}
    \label{tab:data details}
\end{table*}

The foundational information pertaining to the datasets employed in our experiments is delineated in Table~\ref{tab:data details}. In order to ensure equitable comparisons across different methodologies, we adhere to the configuration established in {\tt pFedSim}. This involves the segmentation of each dataset into 100 subsets, effectively constituting 100 clients. It is essential to note that the data distribution throughout the experiments is held constant, thus providing a consistent basis for evaluating the efficacy of diverse methods.

\subsubsection{Model Architecture}

LeNet5~\cite{lecun1998gradient} serves as the backbone model in our experiments. The detailed architectures and the two extended architectures are illustrated in Table~\ref{tab:lenet5}-Table~\ref{tab:extend_lenet5_2}, encompassing the layers with abbreviations in the PyTorch style. 

\begin{table}[h]
    \centering
    \begin{tabular}{p{1.8cm}<{\centering}c}
        \toprule
         Component & Layer \\
        \midrule
         \multirow{6}*{\makecell[c]{Feature\\ Extractor}} & Conv2D (\textit{in}=3, \textit{out}=6, \textit{kernel}=5, \textit{stride}=1, \textit{pad}=0) \\
         & Conv2D (\textit{in}=3, \textit{out}=16, \textit{kernel}=5, \textit{stride}=1, \textit{pad}=0) \\
         & MaxPool2D (\textit{kernel}=2, \textit{stride}=2) \\
         & Flatten() \\
         & FC (\textit{out}=120) \\
         & FC (\textit{out}=84) \\
         \midrule
         Classifier & FC (\textit{out=num\_classes}) \\
        \bottomrule
    \end{tabular}
    \caption{LeNet5 Architecture. Conv2D() consists of a 2D convolution layer, batch normalization layer, and ReLU activation, which is executed sequentially. The \textit{in, out, kernel, stride, pad} represent the input channel, output channel, kernel size, convolution kernel step size moving, padding size, respectively; MaxPool2D() is a max pooling layer for 2D input; Flatten() is for reshaping input from 2D to 1D; FC() is a fully connected layer, out means the number of features output has.}
    \label{tab:lenet5}
\end{table}

\begin{table}[h]
    \centering
    \begin{tabular}{p{1.8cm}<{\centering}c}
        \toprule
         Component & Layer \\
        \midrule
        \multirow{5}*{\makecell[c]{Feature\\ Extractor}} & Conv2D (\textit{in}=3, \textit{out}=6, \textit{kernel}=5, \textit{stride}=1, \textit{pad}=0) \\
         & Conv2D (\textit{in}=3, \textit{out}=16, \textit{kernel}=5, \textit{stride}=1, \textit{pad}=0) \\
         & MaxPool2D (\textit{kernel}=2, \textit{stride}=2) \\
         & Flatten() \\
         & FC (\textit{out}=120) \\
         \hline
         Classifier & FC (\textit{out=num\_classes}) \\
        \bottomrule
    \end{tabular}
    \caption{Extended LeNet5 Architecture 1. Conv2D() consists of a 2D convolution layer, batch normalization layer, and ReLU activation, which is executed sequentially. The \textit{in, out, kernel, stride, pad} represent the input channel, output channel, kernel size, convolution kernel step size moving, padding size, respectively; MaxPool2D() is a max pooling layer for 2D input; Flatten() is for reshaping input from 2D to 1D; FC() is a fully connected layer, out means the number of features output has.}
    \label{tab:extend_lenet5_1}
\end{table}

\begin{table}
    \centering
    \begin{tabular}{p{1.8cm}<{\centering}c}
        \toprule
         Component & Layer \\
        \midrule
        \multirow{7}*{\makecell[c]{Feature\\ Extractor}} & Conv2D (\textit{in}=3, \textit{out}=6, \textit{kernel}=5, \textit{stride}=1, \textit{pad}=0) \\
         & Conv2D (\textit{in}=3, \textit{out}=16, \textit{kernel}=5, \textit{stride}=1, \textit{pad}=0) \\
         & MaxPool2D (\textit{kernel}=2, \textit{stride}=2) \\
         & Flatten() \\
         & FC (\textit{out}=120) \\
         & FC (\textit{out}=84) \\
         & FC (\textit{out}=64) \\
         \hline
         Classifier & FC (\textit{out=num\_classes}) \\
        \bottomrule
    \end{tabular}
    \caption{Extended LeNet5 Architecture 2. Conv2D() consists of a 2D convolution layer, batch normalization layer, and ReLU activation, which is executed sequentially. The \textit{in, out, kernel, stride, pad} represent the input channel, output channel, kernel size, convolution kernel step size moving, padding size, respectively; MaxPool2D() is a max pooling layer for 2D input; Flatten() is for reshaping input from 2D to 1D; FC() is a fully connected layer, out means the number of features output has.}
    \label{tab:extend_lenet5_2}
\end{table}

\subsubsection{Full Hyperparameter settings}
We provid a comprehensive list of hyperparameter settings for all the comparative methods mentioned above. The majority of baseline hyperparameters align with the values specified in their own papers.
\begin{itemize}
    \item {\tt FedProx}~\cite{li2020federated} We set $\mu =1$.
    \item {\tt FedRep}~\cite{collins2023exploiting} We set the epoch for training feature extractor part to 1.
    \item {\tt pFedSim}~\cite{tan2023pfedsim} We set the generalization ratio $\rho_{general}=0.5$.
    \item {\tt FedAP}~\cite{lu2022personalized} We set model momentum $\mu = 0.5$. 
    \item {\tt FedDyn}~\cite{acar2021federated} We set $\alpha = 0.01$.
    \item {\tt FedFomo}~\cite{zhang2020personalized} We set the \textit{M} = 5 and the ratio of validation set to 0.2.
    \item {\tt FedCMD} (Ours) We set the devision ratio $\rho=0.1$.
\end{itemize}

\end{document}